\documentclass{article}

\usepackage[preprint]{neurips_2026}

\usepackage{amsmath}
\usepackage{amssymb}
\usepackage{hyperref}
\usepackage{listings}
\usepackage{svg}
\graphicspath{{figures/}}
\usepackage[labelfont=bf]{caption}
\usepackage{subcaption}
\usepackage{tabularx} 
\usepackage{algorithm}
\usepackage{algpseudocode}

\usepackage[utf8]{inputenc} 
\usepackage[T1]{fontenc}    
\usepackage{hyperref}       
\usepackage{url}            
\usepackage{booktabs}       
\usepackage{amsfonts}       
\usepackage{nicefrac}       
\usepackage{microtype}      
\usepackage{xcolor}         

\title{Text Corpora as Concept Fields: Providing Explainable Groundedness}

\author{
  Nicholas S. Kersting\\ 
  nicholas.kersting@oracle.com \\
  Oracle Corporation
  \And
  Vittorio Castelli \\
  vittorio.castelli@oracle.com \\
   Oracle Corporation
  \And
  Chieh Ting Yeh \\
  jay.yeh@oracle.com \\
     Oracle Corporation
   \And
  Xinzhu Wang\\
  xinzhu.wang@oracle.com \\
     Oracle Corporation
      \And
  Saad Taame \\
  saad.taame@oracle.com \\
     Oracle Corporation
       \And
  Khaoula Allak \\
  khaoula.allak@oracle.com \\
     Oracle Corporation
}

\begin{document}

\maketitle

\begin{abstract}
 We introduce the \textbf{Concept Field} of a text corpus: a local drift field with pointwise uncertainty, estimated in sentence-embedding space from the deltas between consecutive sentences. Given a candidate sentence transition, we score its agreement with the field by $\zeta$, the mean absolute z-distance between the observed delta and the field's local statistical estimate, giving a quantitative measure of groundedness. The score is model agnostic and corpus-attributable, providing traces to corpus semantics either supporting or refuting groundedness. We drive the computation with the introduction of a \textbf{Vector Sequence Database (VSDB)} that stores embeddings together with sequence-position and next-delta metadata. We demonstrate this approach on two large-scale settings: groundedness detection over the U.S. Code of Federal Regulations, and ungroundedness (novelty) detection over Project Gutenberg, where we show Concept Fields achieve strong selective classification performance under a grounded / ungrounded / unsure triage policy. Unlike retrieval-centric baselines, the resulting coverage-risk behavior shows similar empirical coverage-performance trends in the two evaluated domains. Concept Fields provide a fast, lightweight, and interpretable signal for groundedness, complementary to LLM-as-judge and white-box detectors.
\end{abstract}

\section{Motivation and Introduction}

Whether the output of an LLM is grounded in a knowledge corpus or not is central to its applicability. In cases where factuality is important, one wants to avoid confabulations, precisely where a lack of groundedness with respect to some reference knowledge occurs. The persistent occurrence of confabulations remains one of the main problems with applying LLMs to widely-applicable societal use cases. For applications where creativity or novelty is important, on the other hand forming another central use case of LLMs, one discourages groundedness with respect to existing literature at the very least to avoid inadvertent plagiarism. 

\begin{figure}[t]
    \centering
		\includegraphics[width=0.9\columnwidth,height=0.7\textheight,keepaspectratio]{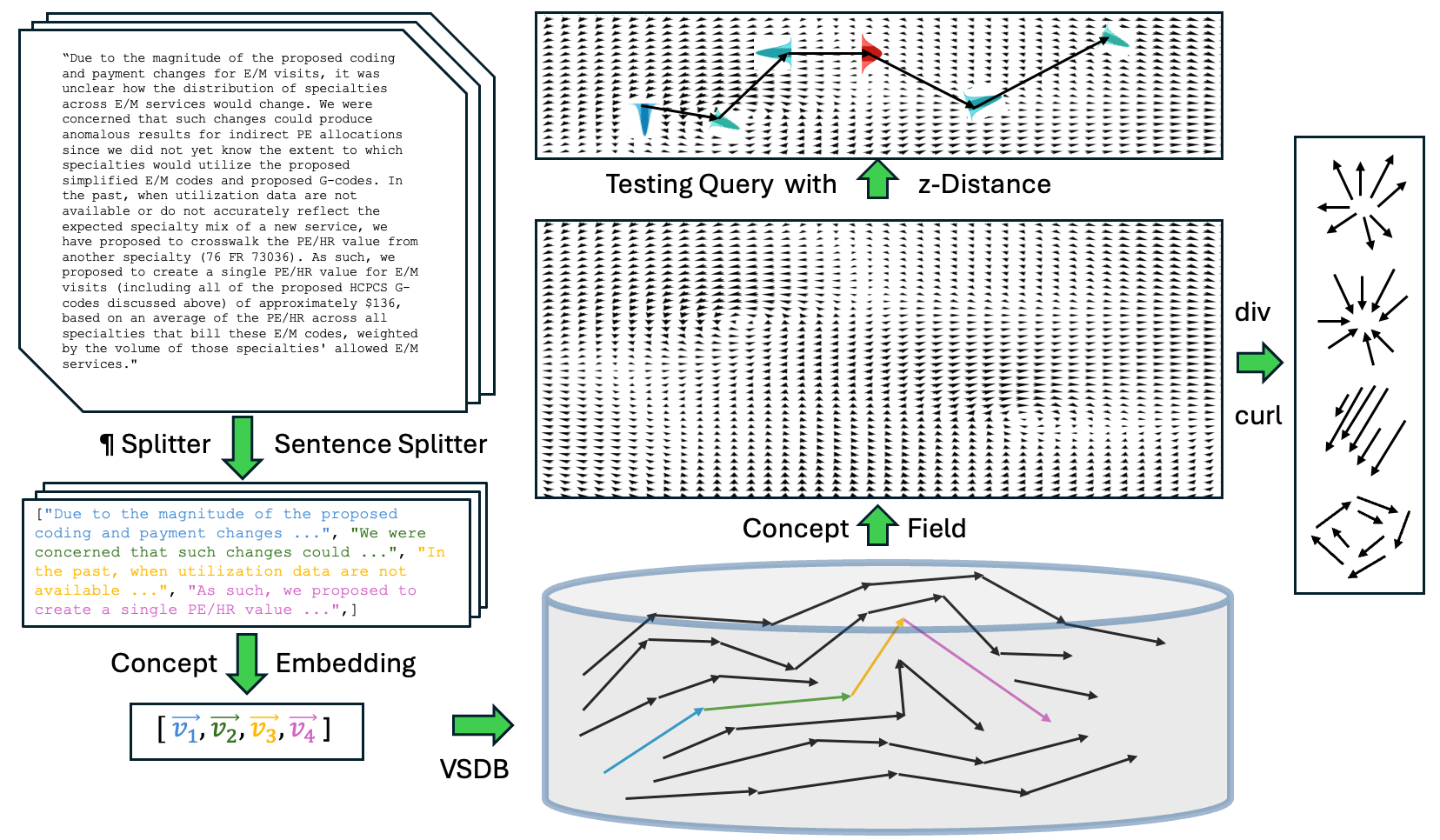}
        \caption{General design: raw text is split into sentence sequences, embedded as concept vectors, and stored in a VSDB. The induced Concept Field supports groundedness scoring by mean absolute z-distance, explainability by nearest corpus sequences, and also exposes whole-corpus geometric features.}
        \label{fig:arch}
\end{figure}

Our approach to this problem is geometric: a text corpus, being a collection of sequential textual units $t_i$ chosen from a lexicon $\Sigma$, can be embedded in a d-dimensional space as vectors $v_i$:
\begin{equation}
\{(t_1,\dots,t_n)\mid t_i\in\Sigma\}\to\{(v_1,\dots,v_n)\mid v_i\in\mathbb{R}^d\}.
\end{equation}
Each sequence of length $n$ thus gives $n-1$ deltas $v_{i+1}-v_{i}$ which we can associate with the respective textual units $t_i$: collecting all such associations across the corpus for each textual unit, we obtain ``flow anchors" in embedding space which then allow us to define a vector field at any point as the average flow of the nearest anchors to that point. While choosing units at the token level may make deltas very washed out (e.g., the word `the' may have many disparate deltas to the next word which cancel out), choosing them at the sentence level with a concept embedding model such as SONAR\cite{duquenne2023sonar} may represent a more meaningful flow. We thus define the Concept Field of a text corpus as a vector field which, given a concept in an embedding space, gives the delta to the next concept as statistically supported by the corpus in the way described above. 
Specifically, we may obtain this field at any point by averaging over a weighted distribution of nearest corpus deltas, giving a mean (delta to the next continued concept) and standard deviation summarizing variation (point-wise uncertainty) among the retrieved transitions.  Broad concepts, for example, may carry larger uncertainty in their continuations while narrower concepts are more strongly-constrained. 
As a stand-alone entity, the Concept Field itself already has potentially interesting flow geometry: regions of high divergence, for example, would indicate concepts converging to or diverging from a central theme; `laminar' regions of low divergence and curl would indicate some standardized logic pattern. Figure~\ref{fig:arch} schematically illustrates our setup.

Our main motivation in this work, however, is to test whether a given sequence of sentences (say from an LLM) is grounded with respect to the Concept Field of a given corpus; this means expressing this sequence as concept embeddings and measuring departure from the Concept Field predictions at each point: we will use the component-wise mean absolute z-distance $\zeta$  between the sequence deltas and the vector field values as a proxy of such. As a standardized deviation measure, small $\zeta$ indicates local agreement with corpus flow; large $\zeta$ indicates departure. And since the computation of $\zeta$ arises from known corpus sentences, we can attribute the (un)groundedness to specific pieces of the corpus.

In the following we will describe  the usage of a Concept Field to detect  clearly grounded versus clearly ungrounded samples, introducing a practical storage/indexing layer, the Vector Sequence Database (VSDB), to support this computation at scale. We then demonstrate the method on a pedagogical 2D system and on two large text corpora, one for factual/legal knowledge (CFR\cite{CFR}), and one for general literature (Gutenberg\cite{Gutenberg}).

Our main contributions in this work are as follows:
\begin{enumerate}
\item \textbf{Concept Field via VSDB.} A corpus-level representation of sentence-sequence flow with local uncertainty, implemented through a new data structure, the VSDB.
\item \textbf{Selective groundedness detection.} $\zeta$-based triage into grounded / ungrounded / unsure, with strong error rates on accepted cases.
\item \textbf{Explainable attribution.} Identification of corpus sentences (refuting)supporting a (un)grounded example alongside the Concept Field predicted flow for comparison.
\item \textbf{Cross-domain application.} Demonstration that the same mechanism supports both compliance-style groundedness checks (in the CFR) and novelty-style deviation checks (against the Gutenberg corpus).
\end{enumerate}

\section{Related Work}
\label{sec:related}

Existing methods for groundedness (typically framed as ``hallucination", i.e. confabulation, detection) are either white-box (model-internal)\cite{doan2025maestro,wang2025mirage,zhang2025siren}, grey-box (token probabilities/perplexity)\cite{xue2025verify, factselfcheck2026, farquhar2024detecting, kuhn2023semantic}, or black-box but calibration-heavy (e.g., LLM-as-judge, perturbation stability thresholds, cosine cutoffs)\cite{zheng2023judging, liu2023geval, pradhan2025legalrag, kersting2024harmonicllmstrustworthy}. These approaches can be expensive, architecture-dependent, or difficult to interpret operationally.
Our approach, on the other hand, is fully black-box, explainable, and yields an interpretable standardized deviation metric with a probabilistic motivation. 

Black-box hallucination detection through self-consistency methods such as SelfCheckGPT~\cite{manakul2023selfcheckgpt}, SAC$^3$~\cite{zhang2023sac3}, and FactSelfCheck~\cite{factselfcheck2025} rely on multiple generations. By contrast, our method scores a single generated sequence against an external corpus field. Geometric output-space approaches~\cite{ricco2025embedding,cert2026dc,cert2026taxonomy,geom_uncertainty2025} are closest in spirit; our key difference is corpus-wide multi-step flow with explicit local uncertainty.
Other works consider sequence comparison in embedding space, including DTW-style methods~\cite{sakoe1978dtw,liu2007dtw_sentence,zhu2017dtw_paragraph,adpm_dtw2025} which are fundamentally pairwise. In contrast, our approach compares a query against a field induced by the whole corpus. Chamfer-style matching~\cite{khattab2020colbert,muvera2024} and learned sequence distance models~\cite{gnesda2025} are complementary.

Finally, our work is based on the use of SONAR~\cite{duquenne2023sonar} (and note OmniSONAR~\cite{omnisonar2026}) with FAISS~\cite{douze2024faiss}; field interpolation uses Inverse Distance Weighting (IDW)~\cite{shepard1968idw}.

\section{Vector Sequence Database (VSDB)}
\label{sec:vsdb}

To support computation of a Concept Field, we introduce the VSDB, a construct which stores variable-length sequences of fixed-dimensional vectors
$V_{[1]},V_{[2]},\dots$, with $V_{[k]}=\{v_1,\dots,v_n\},\;v_i\in\mathbb{R}^d$. Examples in d=1 might include stock ticker time series, single-track musical scores, and seismographs.  Spatial trajectories are good examples of 2- or 3-d vector sequences, which provide our pedagogical example in the next section, and  for the remainder of this paper vectors are high-dimensional sentence embeddings and sequences are paragraph-level sentence chains.

Regardless of the specific application, for each stored vector we keep sequence-local metadata:
\begin{equation}
VSDB\equiv\left\{\left(v_i,\left\{(ID_j,p_{i,j},\Delta_{i,j})\mid v_i\in V_{[q]_j}\right\}\right)\right\},
\end{equation}
\begin{equation}
\Delta_{i,j}=\begin{cases}
v_{p+1}-v_p,&p<m\\
\text{null},&p=m
\end{cases}
\end{equation}
where $p$ is the position in sequence $j$ labeled by $ID_j$.

This design supports:
\begin{enumerate}
\item fast nearest-neighbor lookup (we use FAISS\cite{douze2024faiss}),
\item sequence retrieval/reranking (ID frequency + Chamfer cosine-distance between sequences),
\item local delta aggregation (0th-order field estimate summing the deltas for each vector),
\item IDW interpolation for Concept Field estimation from nearby anchors (1st-order field estimate from Euclidean weighting)
\end{enumerate}

We use a minimal implementation with heuristic settings sufficient for this study. IDW interpolation uses inverse $1/d$ weighting, retaining at most $topN$ neighbors within distance $d_{max}$. To compute the z-distance deviation metric of a delta D against the Concept Field value $\mu$, we take a component-wise average over the largest $k=topN_\zeta$ components of their difference, normalized by component-wise standard deviation $\sigma$ of $\mu$:
\begin{equation}
\label{eqn:zeta}
    \zeta(D)
    =
    \frac{1}{k}\sum_{i=1}^{k}
    \left|
    \frac{D_i - \mu_i}{\sigma_i}
    \right|.
\end{equation}
This coordinate selection is a heuristic intended
to reduce sensitivity to low-magnitude directions
in high-dimensional embedding space.
 We address tuning of all hyperparameters later.

In this work $\zeta$ thus serves as a locally normalized standardized-deviation score, with larger values corresponding to lower likelihood under the local corpus flow. In experiments, we use $\zeta$ both as a continuous risk signal and as the basis for empirical triage thresholds.

\section{Pedagogical Example: 2D Ballistics}
\label{sec:toy}

\begin{figure}[htbp]
  \centering

  \begin{subfigure}[b]{0.47\linewidth}
    \centering
    \includegraphics[width=\linewidth,height=0.22\textheight,keepaspectratio]{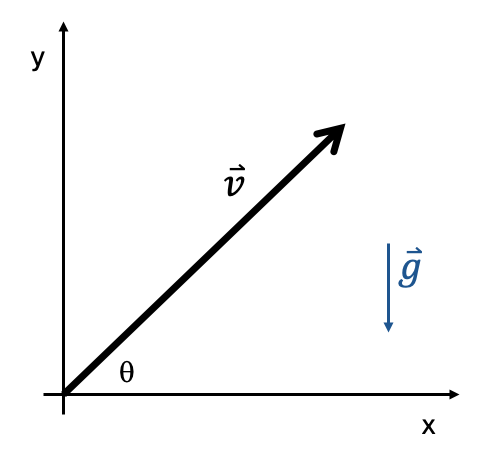}
    \caption{Projectile setup with launch angle $\theta$.}
    \label{fig:physics}
  \end{subfigure}\hfill%
  \begin{subfigure}[b]{0.47\linewidth}
    \centering
    \includegraphics[width=\linewidth,height=0.22\textheight,keepaspectratio]{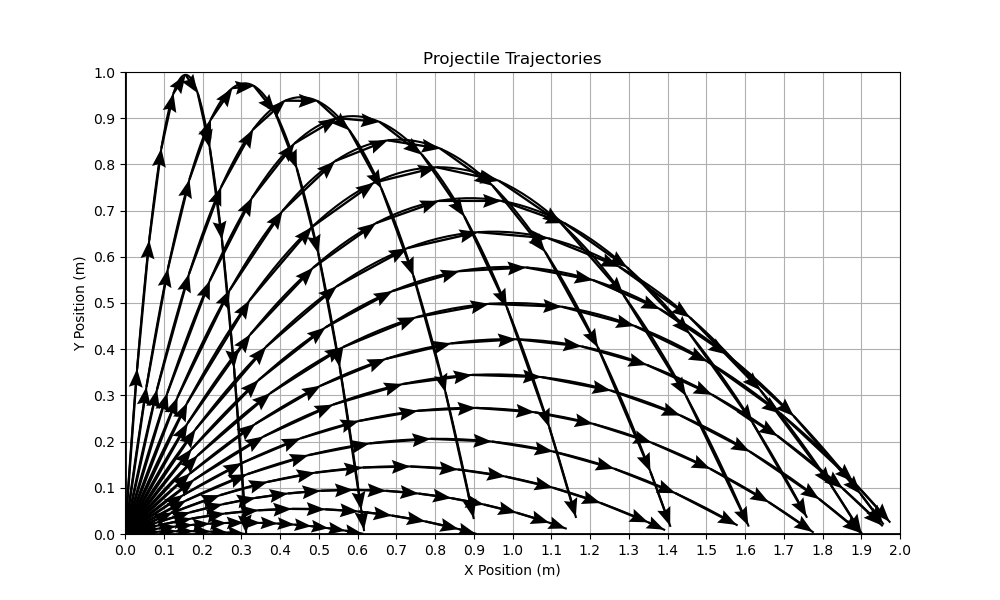}
    \caption{Some of the trajectories used to construct the VSDB.}
    \label{fig:all_traj}
  \end{subfigure}\par

  \begin{subfigure}[b]{0.47\linewidth}
    \centering
    \includegraphics[width=\linewidth,height=0.22\textheight,keepaspectratio]{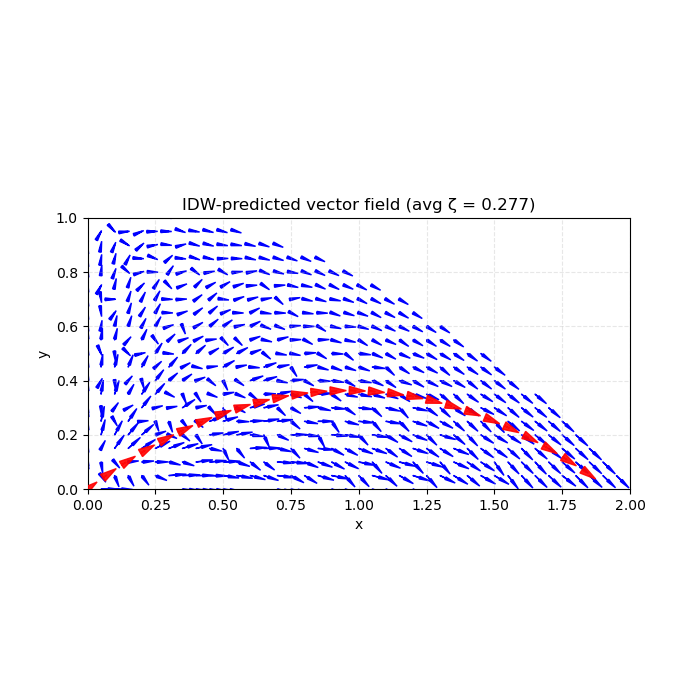}
    \caption{IDW field comparison with matching physics.}
    \label{fig:real_traj}
  \end{subfigure}\hfill%
  \begin{subfigure}[b]{0.47\linewidth}
    \centering
    \includegraphics[width=\linewidth,height=0.22\textheight,keepaspectratio]{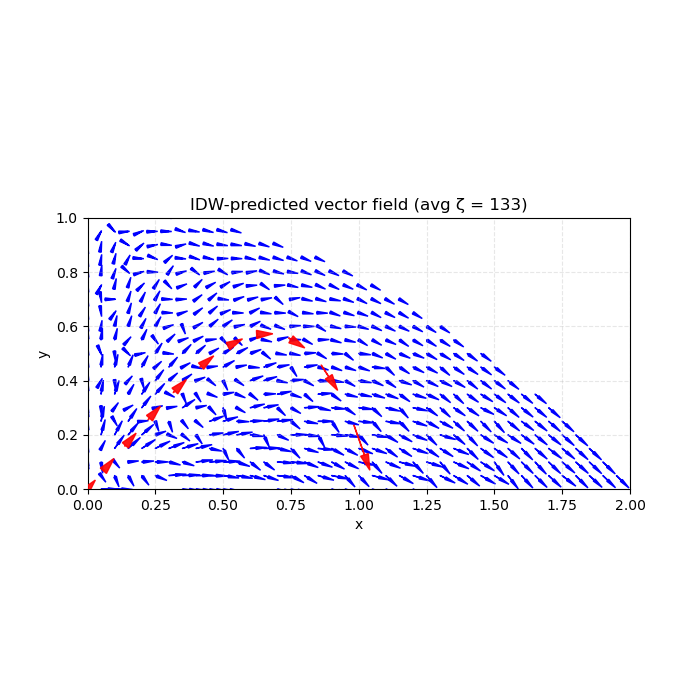}
    \caption{Comparison with air resistance.}
    \label{fig:air_resist}
  \end{subfigure}\par

  \begin{subfigure}[b]{0.47\linewidth}
    \centering
    \includegraphics[width=\linewidth,height=0.22\textheight,keepaspectratio]{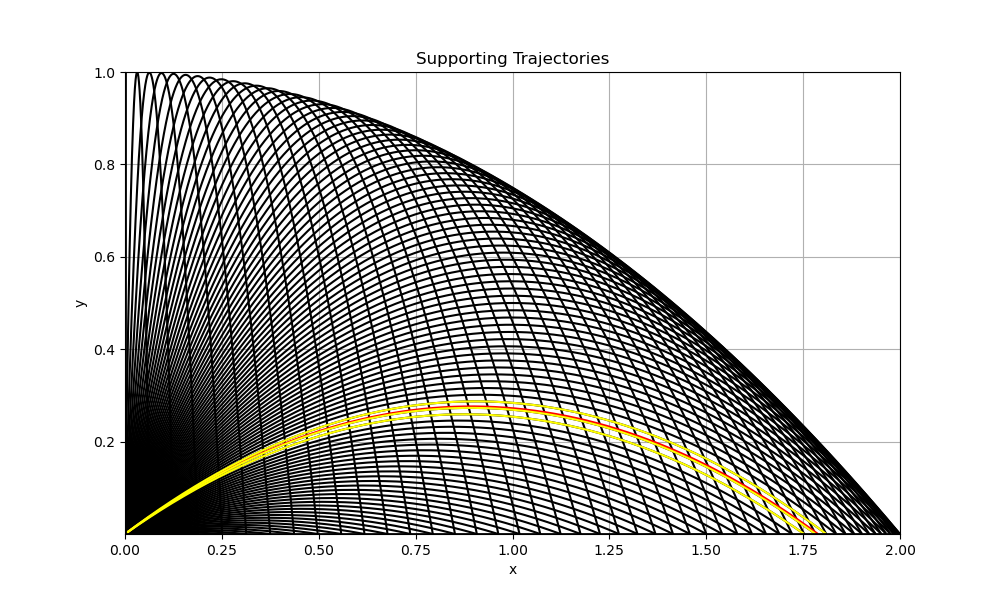}
    \caption{The closest corpus trajectories (yellow) to the grounded example (red) `support' it as an interpolation.}
    \label{fig:support1}
  \end{subfigure}\hfill%
  \begin{subfigure}[b]{0.47\linewidth}
    \centering
    \includegraphics[width=\linewidth,height=0.22\textheight,keepaspectratio]{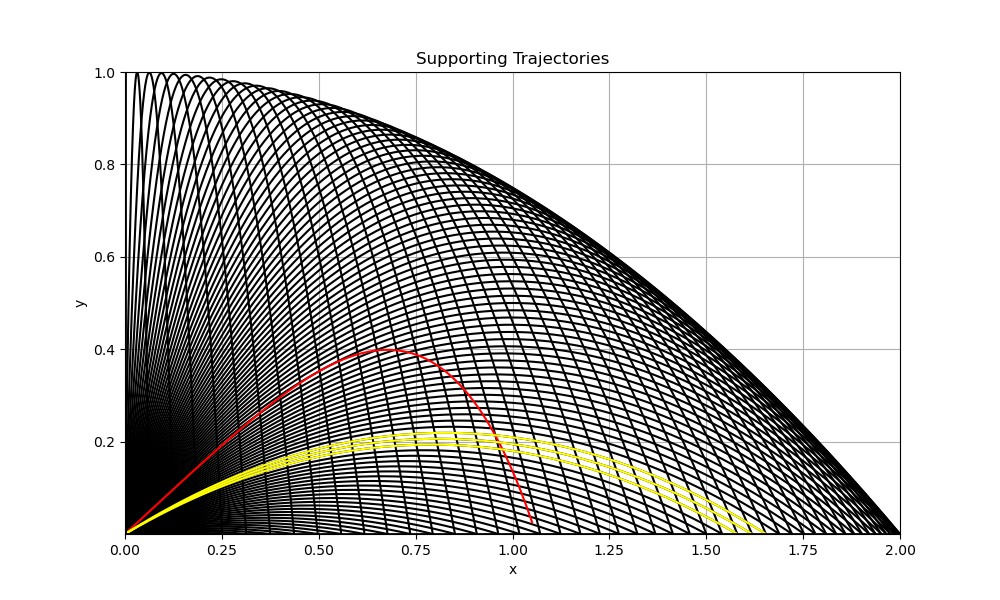}
    \caption{For an ungrounded example (red), the closest trajectories (yellow) do not provide support.}
    \label{fig:support2}
  \end{subfigure}\par

  \caption{Illustration of a Concept Field for 2D ballistic trajectories.}
  \label{fig:single_column_3x2}
\end{figure}

The principles and implementation details of computing a Concept Field from a VSDB are best illustrated in a low dimensional intuitive example. Here we consider the textbook physics treatment of ballistic trajectories in 2D with no air resistance and constant gravity pulling downwards (Figure~\ref{fig:physics}). If we imagine launching many such projectiles from the origin with the same initial speed but at launch angles varying uniformly from 0 (i.e., just skimming the ground) to 90 (straight upwards), we obtain a series of parabolic trajectories like those shown in Figure~\ref{fig:all_traj}; it is then straightforward to discretize (in time or position) and store these as 2D vector sequences $[(0,0), (x_1, y_1), ... , (x_n, y_n)]$ in a VSDB.

Now we can compute the Concept Field of this VSDB at any point by averaging the first derivatives of the $topN$ closest trajectories and weighting by inverse distance; performing this on a grid we obtain a sampling of the vector field shown by blue arrows in Figure~\ref{fig:real_traj} (precise parameters of this setup are described in Appendix C). Here the `concepts' are just spatial points and the Concept Field is describing the most likely path proceeding from any given point, given that all trajectories in the `corpus' are equally likely.
Note that near the origin, where many trajectories cross, the vector field rapidly changes in a noisy fashion, a hallmark of the inherent uncertainty in that region. This uncertainty fades away towards the edge of the kinematic envelope (also a parabola) where all trajectories tend to flow together: the most likely path is quite certain there. 

Figure~\ref{fig:real_traj} also shows sampled velocity vectors of a query trajectory (the red arrows) along its path, and we can compare these with the underlying vector field's values, using Eqn.~\ref{eqn:zeta} (with $k=2$) to test statistical agreement. We compute $\zeta = 0.277$, giving a small standardized deviation from the estimated field. Note here we do not attain perfect agreement due to the irreducible  uncertainty in the trajectory passing through the inherently noisy region near the origin, and we can take it as `good enough' agreement with the underlying vector field (it is common practice in experimental physics, for  example, to take exclusion thresholds closer to $3\sigma$ or higher). This would be an example of a 'grounded' trajectory.

Now for a query trajectory generated with additional physics (we add air resistance) in Figure~\ref{fig:air_resist}, we see the trajectory departs from a parabolic shape and thus expect statistical disagreement; indeed we measure $\zeta = 133$, giving an extremely large standardized deviation from the estimated field: the fact that different physics is involved correctly shows up as a signal of ungroundedness.

Note also that for a grounded trajectory we find that the closest (by Euclidean measure) corpus examples straddle it to either side and `explain' it in the sense of interpolation (Figure~\ref{fig:support1}).
For an ungrounded example, however, the closest trajectories do not give such support (Figure~\ref{fig:support2}).

The usefulness of this simple example is to provide a direct analogy to the more complex situation with higher-dimensional text embeddings: in constructing a VSDB from sequences of natural language sentence embeddings, we will encounter the same use of the IDW vector field and its geometrical features, with $\zeta$ of a query sequence providing a proxy for groundedness with respect to the corpus Gestalt, and closest `trajectories' providing explanatory support.

\section{The Concept Field of a Text Corpus}
\label{sec:manifold}

The construction of a high-dimensional text corpus Concept Field will proceed in a similar manner to the 2D example above with several additional complications: now the `trajectories' consist of sequences of English sentences, possibly of varying length, contained in paragraphs, requiring sentence and paragraph splitting; such trajectories are very sparsely distributed in 1024-d space; the IDW field we considered cannot be realistically constructed on a grid but rather more appropriately on a point-by-point basis: it may be highly irregular and suffer low statistical significance in most of space. Despite these caveats, we will show the procedure still makes sense and yields practical results.

\begin{figure}[ht]
    \centering
        \includegraphics[width=\linewidth,height=0.7\textheight,keepaspectratio]{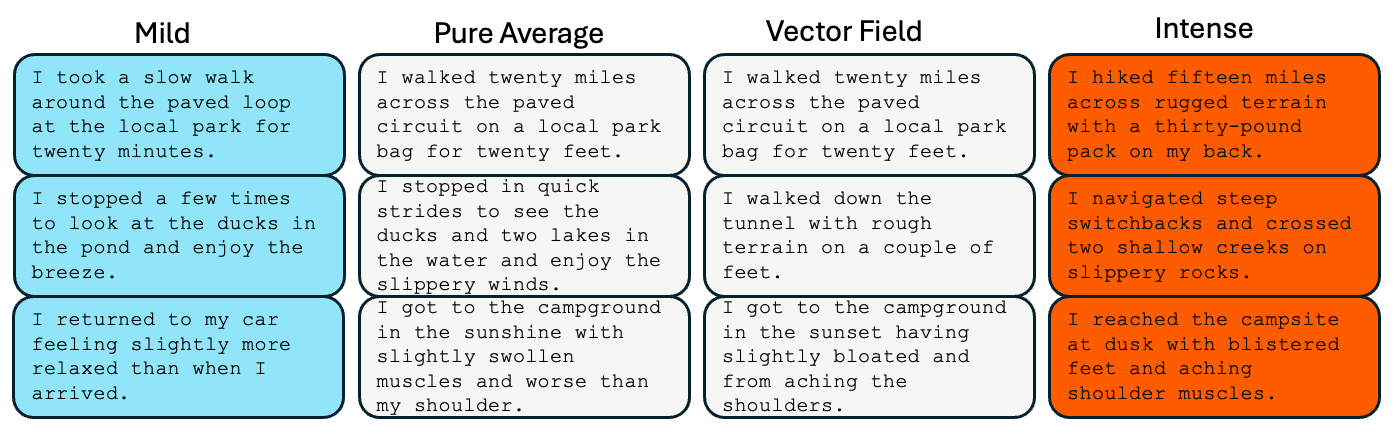}
        \caption{Interpolation check. Left/right columns are mild/intense source sequences; middle columns show SONAR-decoded embedding averages and IDW field drift outputs.}
        \label{fig:sent_interp}
\end{figure}

We first check whether the field interpolates plausibly between nearby sequences. Creating a VSDB from two sentence sequences, one a more intense version of the other,  we compare
the pure average embedding versus the embedding obtained from IDW field drift starting from the same initial point, both decoded into text by SONAR. 
In Figure~\ref{fig:sent_interp} we show one instance of this exercise (the Appendix contains additional ones), and as we see, both produce coherent intermediate semantics, supporting the practical use of both SONAR and local IDW field interpolation.

\begin{figure}[ht]
    \centering
        \includegraphics[width=\linewidth,height=0.7\textheight,keepaspectratio]{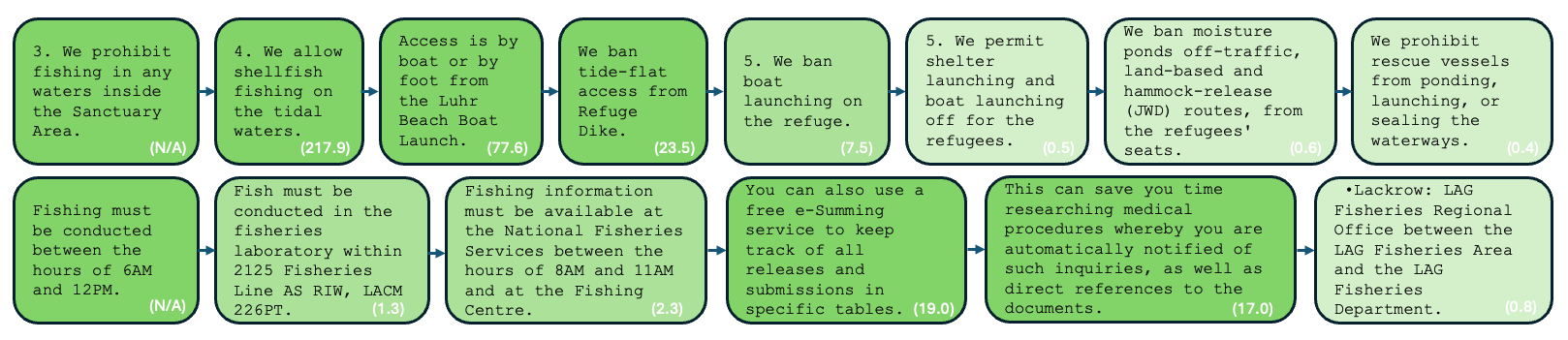}
        \caption{Field-following examples. Top: starting from an in-corpus CFR sentence and tracking a supported drift using the IDW Concept Field. Bottom: starting from an arbitrary sentence and following the local field where defined. Significance (the ratio of drift magnitude to its standard deviation) values are in white.}
        \label{fig:freewalk}
\end{figure}

As another check, on a realistic corpus (CFR 2016) field-following from an in-corpus sentence remains close to the source sequence while local support is strong; once support weakens, significance (defined by the average ratio of the delta to its standard deviation) drops and drift becomes unstable: see Figure~\ref{fig:freewalk}.

\subsection{Testing Groundedness}
\label{subsec:truthfulness}

Now for our main experiments, we evaluate two different uses of $\zeta$ for groundedness:
\begin{enumerate}
\item \textbf{Groundedness in the CFR} (ungrounded = confabulation): we consider publicly-available CFR data from 2000-2025 (comprising about 70M sentences/1B tokens) as a source of U.S. legal truth. Encoding each block of sentences as a vector sequence, we build a VSDB sharded by year and first test whether LLM-reworded pairs\footnote{It is for efficiency's sake that we test pairs of sentences rather than entire sequences. In practice of course the latter are built of the former and one can aggregate $\zeta$.} of neighboring sentences are reliably identified as either faithful to the original meaning or not based on $\zeta$ of these sentence pairs alone. We then perform a confabulation test by generating question-answer pairs from CFR context with a stronger LLM and then asking a weaker LLM to answer without this context, seeing how the correctness of the answer correlates with $\zeta$. 

\item \textbf{Groundedness in Gutenberg} (ungrounded = novelty):  we encode the English language portion of the pre-March 2026 Gutenberg 75k corpus (comprising about 57k books/5B tokens) as a source of literature. Again we test LLM-reworded pairs of neighboring sentences to see if large $\zeta$ reliably identifies the novel aspect of the sentence transitions. As a more practical test of novelty, we then test whether non-English translations of included Gutenberg works (being non-trivially distinct without being conceptually novel) are well separated from more recent, non-included Gutenberg works (by definition novel) based on the $\zeta$ metric alone.
\end{enumerate}

In both settings we thus perform a two-stage test: first with a controlled stress-test using LLM-reworded sentence pairs as data, then with a set-up using $\zeta$ in the way it would probably be used in the wild testing for potential confabulation and novelty. The unifying question is whether $\zeta$ measures groundedness across very different corpora in the same fashion.

\subsubsection{CFR Groundedness (Confabulation Detection)}

As the CFR is a year-by-year corpus, with laws in one year overriding all laws in previous years, we therefore focus on testing laws from individual years. For this purpose, we randomly choose 10 years between 2000 and 2025 as a validation set and the rest for testing.

\begin{figure}[ht]
    \centering
        \includegraphics[width=\linewidth,height=0.7\textheight,keepaspectratio]{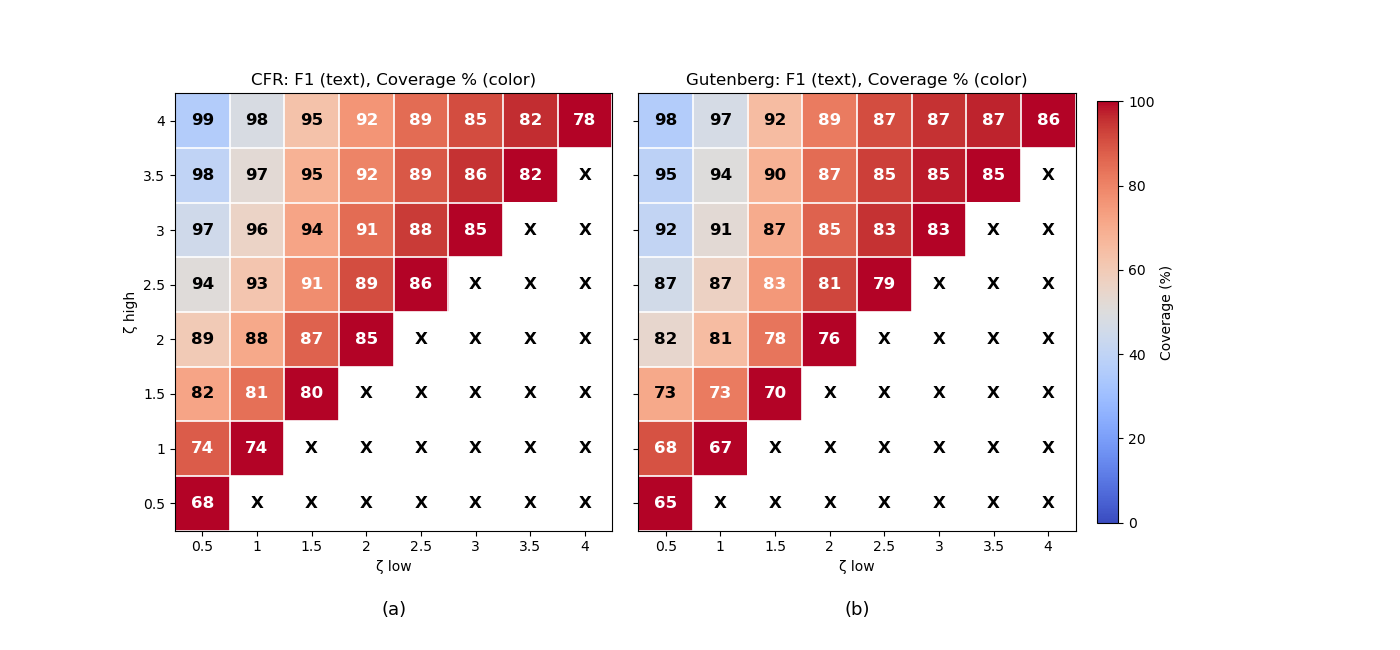}
        \caption{Threshold sensitivity heat maps for (a) Groundedness detection in the CFR, and (b) Novelty detection in the Gutenberg corpus. Despite being very different corpora, the coverage-performance trade-off is similar.}
        \label{fig:heat}
\end{figure}

Now for our first experiment, we use GPT-OSS-120B to rewrite a random selection of 1000 neighboring sentence pairs from each year as either a grounded example (keeping semantics intact) or an ungrounded example (changing semantics significantly), thus forming a balanced 26k+26k set. The Appendix contains details and examples of these transformations, two of which we show in Table~\ref{tab:transf1}.  The same LLM is used to post-filter sentence pairs that don't meet the (un)groundedness criteria and human spot-checking is done at all stages to check the quality of this procedure, which disqualified about 25\% of the generated examples. We then check each sentence pair's delta (i.e., the difference of the embeddings $s_2 - s_1$) against the corpus Concept Field value as measured at $s_1$, expressing the deviation via Eqn.~\ref{eqn:zeta}. Full description of the experimental steps, including tuning of the hyperparameters $(p, ~topN, topN_\zeta,~ d_{max})$ on the 10k+10k Validation Set, is provided in the Appendix, wherein we fix $p=1, ~topN=10, topN_\zeta=50$, and $d_{max}=0.3$.
While the technical usage of $\zeta$ is a standardized measure of departure from locally observed corpus transitions, we also have a practical usage here as a simple classifier of confabulation by setting user-defined thresholds for clearly grounded examples ($\zeta < \zeta_{low}$), ungrounded examples ($\zeta > \zeta_{high}$) and everything else "unsure."  Which values one uses depend on the
intended coverage and error trade-off.
 Figure~\ref{fig:heat}a reports the observed coverage-F1 surface for the CFR controlled test set under the  $[\zeta_{low}, \zeta_{high}]$ unsure interval, where we see that, statistically speaking, $\zeta$ thresholds correlate with confabulation. 
 
 Moreover, since the $\zeta$ computation is backed by a VSDB, we have trajectory-based explainability as well like we saw in our physics example in Figures~\ref{fig:support1},\ref{fig:support2}. Focusing on a high-$\zeta$ example in the first row of Table ~\ref{tab:CFRpredicted-trajectories}, for example, we see that the predicted trajectory by following the Concept Field drift from the first sentence looks quite different with a very high significance, and moreover the closest trajectory in the VSDB likewise conflicts with the example --- these features essentially `explain' the high value of $\zeta$. Likewise, for an example with a low value of $\zeta$ in the second row, the predicted and closest trajectories are quite close in agreement with it, again providing actual attribution for the low $\zeta$ value.

\begin{table*}[t]
  \caption{Examples of Ground Truth sequences from CFR transformed with an LLM to create either a grounded (top) or ungrounded (bottom) example; with respect to the VSDB including the Ground Truth sentence pair, the latter have significantly larger $\zeta$.}
  \label{tab:transf1}
  \centering

  \begin{tabularx}{\linewidth}{
    >{\raggedright\arraybackslash}X
    >{\raggedright\arraybackslash}X
    cc
  }
    \toprule
    \textbf{Ground Truth} &
    \textbf{Transformed} &
    $\zeta_{GT}$ &
    $\zeta_{transf}$ \\
    \midrule

    A discussion of work to be performed during the next reporting period.
    \par\smallskip
    $\longrightarrow$
    \par\smallskip
    Between scheduled reporting dates the grantee(s) also shall immediately inform the Grant Officer's Technical Representative (GOTR) of significant developments affecting their ability to accomplish the work.
    &
    A discussion of work to be performed during the next reporting period.
    \par\smallskip
    $\longrightarrow$
    \par\smallskip
    Between scheduled reporting dates\textbf{,} the grantee(s) \textbf{must also promptly notify} the Grant Officer's Technical Representative (GOTR) of \textbf{any} significant developments \textbf{that affect} their ability to \textbf{complete} the work.
    &
    $4.66 \cdot 10^{-6}$ &
    $1.04$ \\

    \midrule

    Because the Exchange administers the Pilot Fee Structure as part of its rules, the Commission requests that the Exchange provide within 45 calendar days a thorough description of each fee that is permissible under the Pilot Fee Structure.
    \par\smallskip
    $\longrightarrow$
    \par\smallskip
    The description should clearly identify the circumstances in which a distribution intermediary may assess a particular fee.
    &
    Because the Exchange administers the Pilot Fee Structure as part of its rules, the Commission requests that the Exchange provide within 45 calendar days a thorough description of each fee that is permissible under the Pilot Fee Structure.
    \par\smallskip
    $\longrightarrow$
    \par\smallskip
    The description \textbf{must detail the mysterious, almost mythical} circumstances \textbf{that allow} a distribution intermediary \textbf{to conjure and impose a} fee \textbf{at will.}
    &
    $4.49 \cdot 10^{-8}$ &
    $6.14$ \\

    \bottomrule
  \end{tabularx}
\end{table*}

 \begin{table*}[t]
  \caption{The Predicted and Closest trajectories in the 2016 CFR 'explain' the high (first row) or low (second row) values of $\zeta$ for the given test trajectory.}
  \label{tab:CFRpredicted-trajectories}
  \centering
  \small

  \begin{tabularx}{\linewidth}{
    >{\raggedright\arraybackslash}X
    c
    >{\raggedright\arraybackslash}X
    c
    >{\raggedright\arraybackslash}X
  }
    \toprule
    \textbf{Test Trajectory} &
    $\zeta$ &
    \textbf{Predicted Trajectory} &
    \textbf{Significance} &
    \textbf{Closest Trajectory} \\
    \midrule

    Across Highway 68, there are residences along the lake shore.
    $\rightarrow$
    Do not attempt to enter the residences
    &
    $6.44$
    &
    Across Highway 68, there are residences along the lake shore.
    $\rightarrow$
    Utah Lake is a popular area for recreationalists, boaters, and anglers.
    &
    $1.12 \cdot 10^{4}$
    &
    Across Highway 68, there are residences along the lake shore.
    $\rightarrow$
    Utah Lake is a popular area for recreationists, boaters, and anglers.
    $\rightarrow$
    The Lake Mountains are comprised of a
    $\rightarrow$
    The target shooting closure is necessary to protect persons, property, and the public lands and resources in the area.
    $\rightarrow$
    An existing target shooting closure of the area will expire on December 15, 2016 (see 79 FR 74111, December 15, 2014).
    \\

    \midrule

    Across Highway 68, there are residences along the lake shore.
    $\rightarrow$
    Utah Lake is a popular area for enthusiasts, boaters, and fishers.
    &
    $1.65$
    &
   (same as above)
    &
       (same as above)

    &
       (same as above)

    \\

    \bottomrule
  \end{tabularx}
\end{table*}

With the groundwork of the first experiment completed, our second experiment with the CFR testing actual LLM confabulation proceeds straightforwardly: sampling 1000 sequences from the 2016 CFR, we asked GPT-OSS-120B to generate one factual question plus a grounded answer using only the sequence context. The same question without context was then given to GPT-OSS-20B, a much weaker model more prone to confabulation,  to produce an answer which the stronger 120B model judged as either ``supported, partially supported, unsupported, or contradicted," and additionally allowing it to discount the question/answer pair entirely, which eliminated 21 pairs (human quality-checking was performed at all stages of this procedure). We then scored these question-answer pairs with $\zeta$ referencing a VSDB built on this same year of the CFR. Table ~\ref{tab:cfr-20b-zeta-labels} shows that the mean $\zeta$ increases monotonically as support deteriorates, from 1.23 for supported answers to 1.58 for contradicted answers. The contradicted
class therefore has a 29\% higher mean and a 31\% higher median than the supported class, giving credence to using $\zeta$ as a separation metric. For a binary confabulation evaluation, we treated ``supported" answers as negative and combined ``unsupported" and ``contradicted" answers as positive confabulations, excluding the ambiguous ``partially" supported class from this analysis, leaving 829 scored examples: 115 negatives and 714 positives. With $(\zeta_{low}=0.85, \zeta_{high}=1.55)$-thresholds for 50\% coverage, we obtain confabulation detection performance at the $P, R, F1 = 92\%, 78\%, 84\%$ level which is consistent with the prediction of $F1 \sim 81-89\%$ for these thresholds in Figure~\ref{fig:heat}a.

\begin{table*}[t]
  \caption{GPT-OSS-20B answers grouped by the support label assigned
  by GPT-OSS-120B with CFR context.  ``Scored'' excludes rows with
  null $\zeta$ and LLM-as-judge failures.}
  \label{tab:cfr-20b-zeta-labels}
  \centering

  \begin{tabularx}{\linewidth}{>{\raggedright\arraybackslash}Xrrrr}
\toprule
Label & Judged & Scored & Mean $\zeta$ & Median $\zeta$ \\
\midrule
{}Supported & 116 & 115 & 1.23 & 1.06 \\
{}Partially supported & 143 & 142 & 1.35 & 1.30 \\
{}Unsupported & 164 & 163 & 1.42 & 1.39 \\
{}Contradicted & 556 & 551 & 1.58 & 1.40 \\
\bottomrule
\end{tabularx}

\end{table*}

\subsubsection{Gutenberg Groundedness (Novelty Detection)}
A very similar analysis applies to this use case, with however, ungroundedness being a `good' thing: a sentence pair with a Positive label (for novelty) should thus have a large $\zeta$ with respect to the corpus field (or even be `out of corpus', i.e., having no nearby corpus anchors to support a $\zeta$ measurement). Of the roughly 57k English works in the Gutenberg data we collected up to March 2026, we use half (28k) for validation and the rest for testing. The Appendix again contains more details, but we note that  the hyperparameters of our setup optimize to the same values used for the CFR; this, together with the very similar coverage/F1 tradeoff heatmap in Figure~\ref{fig:heat}a-b, demonstrates the feasibility of applying the same scoring mechanism to these two contrasting corpora.

Explainability likewise follows suit: in the first row of Table ~\ref{tab:Gutpredicted-trajectories} we again compare a contrived (novel) Test Trajectory to its predicted and closest trajectories in the Gutenberg corpus, which of course deviate markedly from it, supporting the large value of $\zeta$. The second row shows a Test Trajectory with low $\zeta$, and here the predicted and closest trajectories have semantics essentially conveying the same idea.

 \begin{table*}[t]
  \caption{The Predicted and Closest trajectories in the Gutenberg corpus 'explain' the high (first row) or low (second row) values of $\zeta$ for the given test trajectory.}
  \label{tab:Gutpredicted-trajectories}
  \centering
  \small

  \begin{tabularx}{\linewidth}{
    >{\raggedright\arraybackslash}X
    c
    >{\raggedright\arraybackslash}X
    c
    >{\raggedright\arraybackslash}X
  }
    \toprule
    \textbf{Test Trajectory} &
    $\zeta$ &
    \textbf{Predicted Trajectory} &
    \textbf{Significance} &
    \textbf{Closest Trajectory} \\
    \midrule

    In returning now to the adventures of Siegfried there is little more to be described except the finale of an opera.
    $\rightarrow$
    The finale is entirely predictable as a repetition of the Liebe dynamics of the first act
    &
    $10.22$
    &
    In returning now to the adventures of Siegfried there is little more to be described except the finale of an opera.
    $\rightarrow$
Liefgren, having passed through the fire unscathed, awakens Brindsch and passes through all love's hilarities and nostalgies in a first act at a rehearsal that turns to "Luchendorf Liebe, Liebe der Todt!", which has been eloquently translated as "Love Enlightening to Death, Loving Death", while actually describing loving love and dying love as being closely entwined together as one and the same thing.
    &
    $10.88$
    &
   
In returning now to the adventures of Siegfried there is little more to be described except the finale of an opera.
    $\rightarrow$
Siegfried, having passed unharmed through the fire, wakes Brynhild and goes through all the fancies and ecstasies of love at first sight in a duet which ends with an apostrophe to ’leuchtende Liebe, lachender Tod!’, which has been romantically translated into ’Love that illumines, laughing at Death,’ whereas it really identifies enlightening love and laughing death as involving each other so closely as to be usually one and the same thing."    
    \\

    \midrule

In returning now to the adventures of Siegfried there is little more to be described except the finale of an opera.
    $\rightarrow$
Siegfried, going without harm through fire, awakens Brynhild and falls in love with her, saying something which translates to Love illuminated laughing at Death.    &
    $1.53$
    &
   (same as above)
    &
       (same as above)

    &
       (same as above)

    \\

    \bottomrule
  \end{tabularx}
\end{table*}

Now for our second, practical experiment we compare $\zeta$-values of 1000 sentence pairs collected from 10 post-March-2026 Gutenberg books (thus external to our test VSDB and de facto positive examples of novelty\footnote{These later-release works were verified as absent from the frozen reference index and provide a temporally separated sample for evaluating corpus-relative departure. Gutenberg’s editorial curation motivates the expectation that later additions contain distinct works. }) to 1000 pairs from 10 publicly-available non-English translations of Gutenberg books included in the VSDB, forming a set of non-trivial negative examples (see Table~\ref{tab:books_gutenberg}): concept embeddings are language-agnostic and should thus be very close to those of the original English version, modulo translator discretion. On a per-sentence-pair level, choosing a threshold of $\zeta_{low}=\zeta_{high}=2.1$ for average sentence pair $\zeta$ does indeed identify novel works at the $F1\sim 80\%$ level, in concordance with Figure~\ref{fig:heat}b,  while a more robust book-level identification using the average $\zeta$ is at the $F1 \sim 85\%$ level (Table~\ref{tab:books_gutenberg}). Fuller details of the setup are again in Appendix G.

\begin{table*}[t]
  \caption{Average sentence-pair scores on books Positive(P) or Negative(N) for novelty w.r.t. the Gutenberg corpus; choosing a threshold $\zeta > 2.10$ ($P=81\%, R=90\%, F1 \sim 85\%$) gives comparable performance to a cosine-distance-scored VDB with threshold $d_{cos} >0.494$ ($P=77\%, R=100\%, F1 \sim 87\%$), though the latter is fine-tuned to avoid The First Men in the Moon (Fr).}
  \label{tab:books_gutenberg}
  \centering

  \begin{tabularx}{\linewidth}{>{\raggedright\arraybackslash}Xlrr}
    \toprule
    Book Title & Label & $\zeta$ & $d_{cos}$ \\
    \midrule
    The type-writer girl & P & 2.76 & 0.569 \\
    The adventures of Pony Dexter & P & 2.37 & 0.556 \\
    Magpie---diplomat & P & 2.69 & 0.565 \\
    The lily of Mordaunt & P & 2.51 & 0.558 \\
    The spark of Skeeter Bill & P & 2.25 & 0.532 \\
    The trail of the White Knight & P & 2.20 & 0.571 \\
    The bungalow mystery & P & 2.02 & 0.578 \\
    Out of the depths or, The Rector's trial & P & 2.65 & 0.517 \\
    Such a happy day & P & 2.43 & 0.588 \\
    The burning of Chelsea & P & 2.45 & 0.495 \\
    Père Goriot (Fr) \citep{neg1} & N & 1.98 & 0.376 \\
    The First Men in the Moon (Fr) \citep{neg2} & N & 1.72 & 0.494 \\
    Pride and Prejudice (Fr) \citep{neg3} & N & 2.06 & 0.557 \\
    Illustrious Gaudissart (Fr) \citep{neg4} & N & 1.95 & 0.542 \\
    Pollyanna Grows Up (Ru) \citep{neg5} & N & 1.48 & 0.460 \\
    Ann Veronica (Ru) \citep{neg6} & N & 1.59 & 0.442 \\
    The First Men in the Moon (Ru) \citep{neg7} & N & 1.64 & 0.472 \\
    The First Men in the Moon (Sp) \citep{neg8} & N & 2.31 & 0.300 \\
    Moors and Christians (Sp) \citep{neg9} & N & 2.48 & 0.524 \\
    O Pioneers! (Sp) \citep{neg10} & N & 1.51 & 0.400 \\
    \bottomrule
  \end{tabularx}
\end{table*}

\subsubsection{Baseline Comparisons}

In considering competitive baselines to using the Concept Field, we intentionally discount LLM-based techniques chiefly on the basis of their being inherently untrustworthy, but also typically much more expensive and slower -- an LLM query being at least an order of magnitude slower than a SONAR embedding, the computational bottleneck of the present technique. Comparing only to baselines which are not only corpus-attributable, hence trustworthy, but also equivalently fast and inexpensive to deploy, we settle on a few classifiers applied to the stress-test type experiments of CFR and Gutenberg groundedness detection:
the first two baselines remove the Concept Field and just use the nearest neighbor in the VSDB using either Euclidean (L2) or cosine as distance metrics between the deltas. The third is a vanilla VDB chunked on all pairs of sentences in the corpus, so one is comparing embeddings of sentence pairs with cosine distance to determine groundedness.
Looking at the CFR experiment first in Table~\ref{tab:test_hallucination_cfr},
choosing the "optimal-performance at 50\% coverage" point for comparison, we note that the VDB-based technique generally performs the worst, capturing the smallest number of Positive (confabulation) examples, and tends to err quite heavily on the False Negative side -- this is in keeping with the well-known deficiency of cosine distance to capture negation inside a larger context (see Appendix). VSDB-based techniques, on the other hand, err in the other direction by having slightly more False Positives --- but for risk-averse applications this is actually preferable.
Otherwise the VSDB-based techniques all have comparable performance. However, the Concept Field enjoys three distinct advantages over these other baselines: first, the CFR and Gutenberg operating points occupy similar numerical ranges, with classification thresholds selected according to the desired coverage-error tradeoff in Figure~\ref{fig:heat}a-b; Euclidean L2 or cosine-distance based techniques must always be calibrated on a corpus-by-corpus basis since the scale of these metrics changes with the scope and specific embedding of the corpus data \footnote{We empirically verify that this is the case by computing the corresponding 50\% coverage performances for the Gutenberg Test Set (see Appendix) where the Concept Field thresholds and performance are about the same, while the other baselines vary wildly}. Secondly, unlike top-1 type distance scores, the Concept Field summarizes multiple retrieved continuation vectors through their locally weighted mean and dispersion: $\zeta$ is not just another threshold metric but carries with it the meaning of statistical surprise. And thirdly, the Concept Field provides additional explainability through its ability to predict trajectories (see, e.g., Tables~\ref{tab:CFRpredicted-trajectories},\ref{tab:Gutpredicted-trajectories}) with quantitative significance --- this can be compared to the $\zeta$ value to confirm (un)groundedness.

\begin{table*}[t]
  \caption{Comparison of the Concept Field to various baselines with thresholds (lo,hi) tuned to give maximum performance at 50\% coverage of the original set of CFR 16k+16k pos+neg test samples. Precision is defined on the ungrounded "confabulation" examples. AURC is computed over all possible lo/hi thresholds. Here we see VSDB-based techniques generally outperform a vanilla VDB solution.}
  \label{tab:test_hallucination_cfr}
  \centering

\begin{tabularx}{\linewidth}{>{\raggedright\arraybackslash}p{0.18\linewidth}>{\centering\arraybackslash}X>{\centering\arraybackslash}X>{\centering\arraybackslash}X>{\centering\arraybackslash}X}
\toprule
Measure & Concept Field & VSDB + top1(L2) & VSDB + top1(cos) & VDB + top1(cos) \\
\midrule
{}lo & 1.0 & 0.04 & 0.01 & 0.01 \\
{}hi & 3.75 & 0.15 & 0.14 & 0.13 \\
{}TP & 10479 & 10917 & 7557 & 3566 \\
{}TN & 3974 & 4045 & 7218 & 10600 \\
{}FP & 316 & 293 & 66 & 6 \\
{}FN & 152 & 148 & 566 & 1185 \\
{}P & 0.971 & 0.973 & \textbf{0.991} & \textbf{0.998} \\
{}R & \textbf{0.986} & \textbf{0.987} & 0.930 & 0.751 \\
{}F1 & \textbf{0.978} & \textbf{0.980} & 0.960 & 0.857 \\
{}MCC & \textbf{0.923} & \textbf{0.929} & 0.920 & 0.821 \\
{}AUC & \textbf{0.984} & \textbf{0.986} & 0.973 & 0.919 \\
{}AURC & \textbf{0.039} & 0.037 & \textbf{0.040} & 0.036 \\
\bottomrule
\end{tabularx}

\end{table*}

Now for the more practical confabulation and novelty experiments we discussed above, we compare only to a vanilla VDB built on all sentence pairs using a cosine-distance metric, as this is probably the most readily available and dominant competitor. For the CFR confabulation setup focusing on the 2016 CFR, we retrieved top matching sentence-pairs from the VDB and took cosine distance to the GPT-OSS-20B question-answer pair, fixing coverage at $50\%$ for comparison purposes. We found a threshold of $0.20$ achieving $F1\sim 86\%$, comparable to the Concept Field result.  For the Gutenberg novelty application (see again Appendix G for details) we find that sentence-level novelty detection for the same coverage is at the $F1\sim 74\%$ level at a cosine-distance threshold of $0.42$, significantly below the Concept Field performance; meanwhile book-level detection shoots up to $F1\sim 95\%$ at a very different threshold of $0.494$ (see again Table~\ref{tab:books_gutenberg}) which is even fine-tuned to lie between the very similarly-scored ``The burning of Chelsea" ($d_{cos}=0.495$) and ``The First Men in the Moon (Fr)" ($d_{cos}=0.494$). These results emphasize the fact that cosine-distance based baselines suffer from unpredictable performance levels and threshold values which are hard to determine without explicit calibration to the case at hand, in contrast to our $\zeta$ technique which is fairly stable again as per Figure~\ref{fig:heat}.

\section{Discussion and Conclusions}
\label{sec:discussion}

We have introduced the Concept Field of a text corpus in this work, a conceptually simple yet powerful method to test groundedness with respect to the corpus providing both a probabilistically motivated deviation score and explainability/traceability to the supporting regions of the corpus.  Beyond what we exhibited as  confabulation and novelty detection of LLM outputs, the technique enjoys numerous other applications we did not have space to address, such as  prompt injection defense, corpus-corpus comparison for conflicts, compliance testing, and general corpus assessment based on field curl and divergence. Expanded discussion of the latter appears in the Appendix, and future work will address these in more detail.

As the present default go-to for confabulation detection is LLM-as-judge, we argue that testing groundedness with a Concept Field may become a useful complement or competitor, in particular being a technique that does not depend on the black-box internals of an inherently non-explainable entity. On a practical level as well, a corpus Concept Field is faster (once a VSDB is indexed), cheaper (no API token usage), lightweight (commodity CPU suffices, and one can choose to store the field values separately from the origin sequences, giving lightweight portability),  and enjoys explainability (both by field delta prediction and nearest corpus transitions). In novelty-detection applications where an LLM is of course not applicable (it can't tell you what isn't in its training data), the stability of $\zeta$ as a groundedness metric provides a distinct advantage over the competing cosine-distance metric which suffers from threshold fine-tuning in general. Concepts are also language agnostic, so one corpus Concept Field simultaneously fulfills the same role for testing LLMs specialized on different languages.

Our treatment has been abbreviated and basic by intention, leaving the research community room to improve on it in certain aspects. In particular, one can explore other concept embedding schemes besides SONAR and better interpolation schemes than simple IDW. Some aspects of our approach are fundamentally data-dependent: with more data, the Concept Field fills space more densely, covers more semantic ground, attains more statistical reliability, and the quality of data likewise determines how efficiently this happens:  corpora based on internally consistent content such as Congressional Law fill out embedding space more reliably, while noisier data from the free Internet is probably less efficient. We imagine that as this technique gains favor in the community, people and institutions will cultivate vetted self-consistent corpora and their associated Concept Fields for purposes of testing consistency with curated legal, ethical, or domain-specific reference corpora --- such are useful not only for testing the outputs of LLMs, curbing the ubiquitous confabulation problem, but also any sequence of concepts (even from humans or other corpora). 

Even in this initial form, the approach is practically useful as a complementary trust signal and explainability technique alongside other safeguards, and is applicable beyond text wherever data can be represented as vector sequences.

\bibliography{concept_field}
\bibliographystyle{unsrt}

\section*{APPENDIX}

\subsection*{A. SONAR embeddings}

Clearly our agenda is dependent on the choice of embedding scheme: a poorly-trained embedding will not exhibit meaningful correlations between distance in embedding space and semantics. We therefore explicitly check that SONAR, our choice for this work, is a reasonable scheme.

The first way to check this is to define a linear interpolation Z between sequences X and Y:  $Z = \alpha X + (1-\alpha)Y$, and check semantic stability for $0 < \alpha < 1$.
We do in fact explicitly test across a number of sentence pairs that taking the linear interpolation between the two sentences (concept-embeddings) in each pair we obtain intermediate semantics:

\begin{itemize}
\item X : 'I like to run in the park.'
\item Y :  'I like to walk in the library.'
\item  alpha=0.00: Z: 'I like to run in the park.'
\item  alpha=0.25:  Z: 'I like to run in the park.'
\item  alpha=0.50: Z: 'I like to walk in the park.'
\item  alpha=0.75: Z: 'I like to walk in the library.'
\item  alpha=1.00:  Z: 'I like to walk in the library.'
\end{itemize}

\begin{itemize}
\item  X: 'The food at this restaurant is terrible and the service is slow.'
 \item Y: 'The cuisine at this eatery is excellent and the service is prompt.'
 \item alpha=0.00: Z:'The food in this restaurant is terrible and the service is slow.'
 \item alpha=0.25: Z:'The food in this restaurant is terrible and the service is slow.'
 \item alpha=0.50: Z:'The food in this restaurant is excellent and the service is slow.'
 \item alpha=0.75: Z: 'The cuisine at this eatery is excellent and the service is quick.'
\item  alpha=1.00: Z: 'The cuisine at this eatery is excellent and the service is prompt.'
\end{itemize}

A similar conclusion occurs for generating points randomly in between the two vectors, i.e.,  $Z_i = (X_i + Y_i)/2 + d/2 * \hat{s}$, with $d \equiv |X - Y|$ :
\begin{itemize}
 \item X: 'The meticulous planning for the ambitious cross-continental journey ensured every potential obstacle was identified and systematically mitigated, leading to an utterly smooth and successful outcome.'
 \item Y:  'The haphazard preparation for the challenging cross-continental journey failed to account for several major logistical hurdles, which inevitably resulted in chaotic delays and an ultimate failure of the expedition.'
 \item Z: (1\% closer to Y)  'The meticulous preparation for the pernicious intercontinental journey was not aware of any substantial obstacle arrangements, which were consequently timeless and resulted in the ultimate failure of the expedition.'
\end{itemize}

\subsection*{B. Embedding Perturbation Stability of SONAR (1024-d)}
\label{app:local-gaussian-perturbation}
As a quick sanity check, we perturb the 'delta' between a pair of sentences in a Gaussian fashion and decode with SONAR.

 \noindent\textbf{Original sentence pair}
  \begin{itemize}
  \item \textbf{s1}: ``The annualized and present value estimates of monetized costs and benefits over the 10-year period from 2023 through 2032 using three percent and seven percent discount rates are
  summarized below.''
  \item \textbf{s2}: ``The annualized, monetized costs (2020 USD) of the provisions in the NPRM (if finalized as proposed) are estimated to be \$29 million (range: \$7.7 to \$87 million) using a three percent
  discount rate; the estimated monetized costs using a seven percent discount rate are largely the same.''
  \end{itemize}

  \begin{table*}[t]
  \caption{Noise statistics for 10 perturbations in 1024-dimensional embedding space.}
  \label{tab:perturb_stats_example_full}
  \centering

  \begin{tabularx}{\linewidth}{>{\raggedright\arraybackslash}Xrrrrrr}
\toprule
\textbf{idx} & \textbf{noise1\_l2} & \textbf{noise2\_l2} & \textbf{noise1\_mean} & \textbf{noise2\_mean} & \textbf{noise1\_std} & \textbf{noise2\_std} \\
\midrule
{}1 & 0.03114 & 0.03295 & $-4.92\times10^{-5}$ & $-5.47\times10^{-6}$ & 0.000972 & 0.001030 \\
{}2 & 0.03157 & 0.03202 & $-3.70\times10^{-5}$ & $ 2.71\times10^{-5}$ & 0.000986 & 0.001000 \\
{}3 & 0.03166 & 0.03173 & $ 5.13\times10^{-5}$ & $ 8.61\times10^{-7}$ & 0.000988 & 0.000991 \\
{}4 & 0.03284 & 0.03225 & $ 1.18\times10^{-5}$ & $ 1.52\times10^{-5}$ & 0.001026 & 0.001008 \\
{}5 & 0.03216 & 0.03104 & $ 3.29\times10^{-5}$ & $ 1.49\times10^{-5}$ & 0.001005 & 0.000970 \\
{}6 & 0.03212 & 0.03159 & $-1.64\times10^{-5}$ & $ 2.52\times10^{-5}$ & 0.001004 & 0.000987 \\
{}7 & 0.03134 & 0.03248 & $-1.43\times10^{-5}$ & $ 2.68\times10^{-5}$ & 0.000979 & 0.001014 \\
{}8 & 0.03228 & 0.03077 & $ 2.64\times10^{-5}$ & $-1.74\times10^{-5}$ & 0.001008 & 0.000961 \\
{}9 & 0.03229 & 0.03206 & $-9.69\times10^{-6}$ & $ 4.23\times10^{-5}$ & 0.001009 & 0.001001 \\
{}10 & 0.03111 & 0.03160 & $ 1.18\times10^{-5}$ & $-4.65\times10^{-5}$ & 0.000972 & 0.000986 \\
\bottomrule
\end{tabularx}
  \end{table*}

  \noindent\textbf{Decoded perturbed sentences (all 10)}

  \begin{enumerate}
  \item
  \textbf{s1}: ``The annualized and present value estimates of monetised costs and benefits over the 10-year period from 2023 to 2032 using three per cent and seven per cent discount rates are summarized
  below.''\\
  \textbf{s2}: ``The monetized, annualized costs (2020 dollars) of the provisions in the NPRM (if finalized as proposed) are estimated to be \$29 million (range: \$77--\$87 million) using a three percent
  discount rate; the estimated monetized costs using a seven percent discount rate are largely the same.''

  \item
  \textbf{s1}: ``The annualized and present value estimates of monetised costs and benefits over the 10-year period from 2023 to 2032 using three to 7 per cent discount rates are summarized below.''\\
  \textbf{s2}: ``The annualised, monetized costs (2020 dollars) of the provisions of the NPRM (if finalized as proposed) are estimated to be \$29 million (range: \$77--\$87 million) using a three percent
  discount rate; the estimated monetized costs using a seven percent discount rate are roughly the same.''

  \item
  \textbf{s1}: ``The annualized and present value estimates of monetised costs and benefits over the 10-year period from 2023 to 2032 using three to 7 per cent discount rates are summarized below.''\\
  \textbf{s2}: ``The annualised, monetized costs (2020 dollars) of the provisions of the NPRM (if finalized as proposed) are estimated to be \$29 million (range: \$7.7 million to \$8.7 million) using a three
  percent discount rate; the estimated monetized costs using a seven percent discount rate are roughly the same.''

  \item
  \textbf{s1}: ``The annualized and present value estimates of monetised costs and benefits over the 10-year period from 2023 to 2032 using three per cent and seven per cent discount rates are summarized
  below.''\\
  \textbf{s2}: ``The monetized, annual \$20 million costs (provisions in the NPRM) if finalized (as proposed) are estimated to be \$29 million: \$7.70 million (from \$7.87 million) using a three percent
  discount rate; the estimated monetized costs using a seven percent discount rate are largely the same.''

  \item
  \textbf{s1}: ``The annualized and present value estimates of monetised costs and benefits over the 10-year period from 2023 to 2032 using three per cent and seven per cent discount rates are summarized
  below.''\\
  \textbf{s2}: ``The annualised, monetized costs (2020 dollars) of the provisions of the NPRM (if finalized as proposed) are estimated to be \$29 million (range: \$77--8.7 million) using a three percent
  discount rate; the estimated monetized costs using a seven percent discount rate are roughly the same.''

  \item
  \textbf{s1}: ``The annualized and present value estimates of monetised costs and benefits over the 10-year period from 2023 to 2032 using three per cent and seven per cent discount rates are summarized
  below.''\\
  \textbf{s2}: ``The annualised, monetized costs (2020 dollars) of the provisions of the NPRM (if finalized as proposed) are estimated to be \$29 million (range: \$7.7 million to \$8.7 million) using a three
  percent discount rate; the estimated monetized costs using a seven percent discount rate are largely the same.''

  \item
  \textbf{s1}: ``The annualized and present value estimates of monetised costs and benefits over the 10-year period from 2023 to 2032 using three per cent and seven per cent discount rates are summarized
  below.''\\
  \textbf{s2}: ``The monetized, annual costs (2020 dollars) of the provisions in the NPRM (if finalized as proposed) are estimated to be \$29 million (range: \$7.7 to \$8.7 million) using a three percent
  discount rate; the estimated monetized costs with seven percent discount rate are largely the same.''

  \item
  \textbf{s1}: ``The annualized and present value estimates of monetised costs and benefits over the 10-year period from 2023 to 2032 using three per cent and seven per cent discount rates are summarized
  below.''\\
  \textbf{s2}: ``The annualised, monetized costs (2020 dollars) of the provisions of the NPRM (if finalized as proposed) are estimated to be \$29 million (range: \$77--\$87 million) using a three percent
  discount rate; the estimated monetized costs using a seven percent discount rate are roughly the same.''

  \item
  \textbf{s1}: ``The annualized and present value estimates of monetised costs and benefits over the 10-year period from 2023 to 2032 using three per cent and seven per cent discount rates are summarized
  below.''\\
  \textbf{s2}: ``The annualised, monetized costs (2020 dollars) of the provisions of the NPRM (if finalized as proposed) are estimated to be \$29 million (range: \$7.7 million to \$8.7 million) using a three
  percent discount rate; the estimated monetized costs using a seven percent discount rate are roughly the same.''

  \item
  \textbf{s1}: ``The annualized and present value estimates of monetised costs and benefits over the 10-year period from 2023 to 2032 using three per cent and seven per cent discount rates are summarized
  below.''\\
  \textbf{s2}: ``The annualised, monetized costs (2020 dollars) of the provisions of the NPRM (if finalized as proposed) are estimated to be \$29 million (range: \$7{,}700 to \$8{,}700) using a three percent
  discount rate; the estimated monetized costs using a seven percent discount rate are roughly the same.''
  \end{enumerate}

Similar experiments with several noise levels up to 0.1 likewise yielded expected increasing perturbations of the semantics.

\subsection*{C. Physics Details of the 2D Example}
\label{app:physics}

In this simulation we fix a value of gravity at $g=9.81$ and launch speed $v=\sqrt{2g}$ (MKS units understood in this and the following), confining the projectiles to a region in the x-y plane with $0 <= x <= 2$ and $0 <= y <= 1$
,  regardless of launch angle. We simulate each trajectory with a discrete time-step Euler-method-based on the first-order differential equations $y'(t) = v \sin \theta -gt$ and $x'(t) = v \cos \theta$, defining launch angle $\theta = arctan \frac{y'(0)}{x'(0)}$ and using time steps of $0.001$ which results in a sufficiently long vector sequence for most angles. 
We generate 1000 trajectories with launch angles uniformly spread between 0 and 90 degrees, and use these sequences to compute the underlying
 IDW vector field of the trajectories at any point by averaging over the instantaneous velocities of the $topN$ neighbor points lying on other trajectories. The parameters of this setup include N=10, a nearest-neighbor distance cut-off of $d_{max} = 0.03$, and simple
 inverse $1/d$ weighting, though probably similar values would also work. Sampled on a grid with spacing $0.05$, this gives the blue vector field shown in Figure~\ref{fig:real_traj}.

Although this is just a toy example, one may imagine a realistic scenario in, say, space exploration, where scientists observe trajectories of debris emitted from an exoplanetary volcano and from the shapes of such try to infer geo-atmospheric conditions such as local gravity, air resistance, and wind speed by reference to trajectories from similar volcanoes on Earth.

\subsection*{D. Additional Sentence Interpolations}

\begin{figure}[ht]
    \centering
        \includegraphics[width=\linewidth,height=0.7\textheight,keepaspectratio]{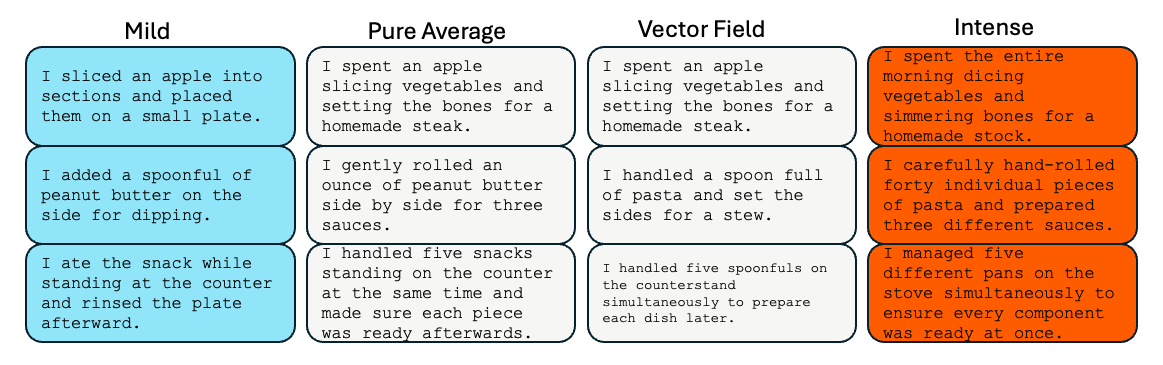}
        \caption{Interpolation check. Left/right columns are mild/intense source sequences; middle columns show SONAR-decoded embedding averages and IDW field drift outputs.}
        \label{fig:sent_interp2}
\end{figure}

\begin{figure}[ht]
    \centering
        \includegraphics[width=\linewidth,height=0.7\textheight,keepaspectratio]{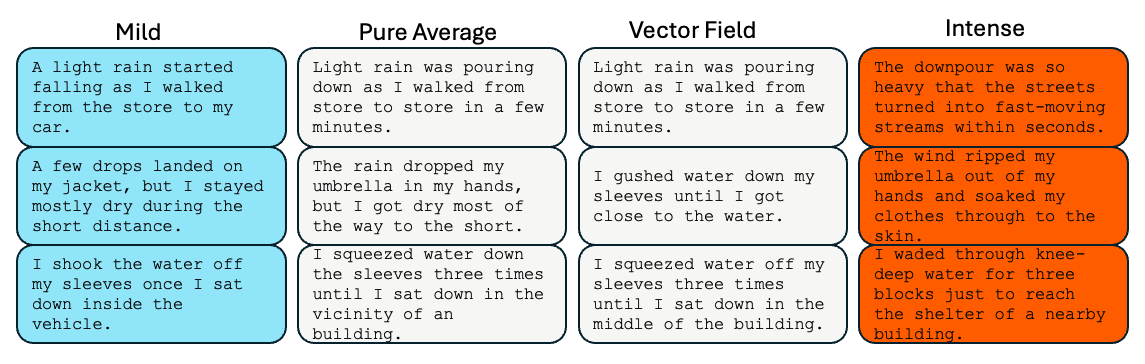}
        \caption{Interpolation check. Left/right columns are mild/intense source sequences; middle columns show SONAR-decoded embedding averages and IDW field drift outputs.}
        \label{fig:sent_interp3}
\end{figure}

\subsection*{E. Fundamental issue with cosine similarity}
Consider the following pair of very similar sentences which in fact only differ by two words:

\begin{itemize}
\item  "However, additional funding may be available in Fiscal Year 2001 to award a limited number of grants under the College and University Affiliations Program to enable current and former Fulbright scholars to build on their experiences as individual Fulbright grantees through broadened institutional cooperation."
\item "However, additional funding may very well be available in Fiscal Year 2001 to award a limited number of grants under the College and University Affiliations Program to enable current and former Fulbright scholars to build on their experiences as individual Fulbright grantees through broadened institutional cooperation."
\end{itemize}

The cosine similarity of their SONAR embeddings is 0.96.
If we negate one of them, 

\begin{itemize}
\item  "However, additional funding may be available in Fiscal Year 2001 to award a limited number of grants under the College and University Affiliations Program to enable current and former Fulbright scholars to build on their experiences as individual Fulbright grantees through broadened institutional cooperation."
\item    "However, additional funding may \textbf{\textit{not}} be available in Fiscal Year 2001 to award a limited number of grants under the College and University Affiliations Program to enable current and former Fulbright scholars to build on their experiences as individual Fulbright grantees through broadened institutional cooperation."
\end{itemize}

the similarity is not only still high, at 0.98, but even higher than the affirmative case!
This illustrates a well-known problem in cosine-similarity measures, especially on small but semantically important changes within larger blocks of text.

\subsection*{F. Groundedness in the CFR}
\label{subsec:truth}

Raw CFR XML-encoded files are available on a year-by-year basis \cite{CFR}. As paragraph boundaries are hard to detect in this corpus, we choose to chunk by every 5 sentences. We use spacy\cite{spacy} for sentence splitting and normalization.

\begin{itemize}
	\item Experimental Procedure

		\begin{itemize}
		   \item Positive examples
		   	 \begin{enumerate}
		   	 	\item extract pairs of sequential sentences from CFR sequences
		   	 	\item reword with LLM to change slightly; use LLM-as-judge with human spot-checking to keep only pairs that do not significantly change semantics;
		   	 		\begin{itemize}
		   	 			\item: Generative Prompt: \textit{``Generate a new second sentence that preserves the meaning of the original second sentence while making only a slight wording change. - Keep the new second sentence very close to the original wording and avoid broad paraphrases, extra details, or a no-op copy.''} Note we keep the first sentence unchanged in order to maximize recall in this limited data set. Relaxing this assumption would simply force us to apply to more data (see $\zeta$ check below, where we discard the sentence pair if its $\zeta$ is too high and `noisy'). 
		   	 			\item: Quality-check Prompt: \textit{``Accept only if:
- the rewritten second sentence is coherent with the first sentence
- it preserves the meaning of the original second sentence
- it makes only a slight wording change rather than a broad paraphrase
- it is not identical to the original second sentence
Reject if any of those are not true.'' }
		   	 		\end{itemize}
		   	 	\item check $\zeta$ of each re-worded pair
		   	 	\item for each re-worded pair, find closest pair (by similarity with first sentence) in VSDB, and its $\zeta$ ; if $\zeta>\zeta_{high}$, consider this pair `too noisy'; i.e., in actual inference application one would not trust this region of the Concept Field, tantamount to the noisy region of the ballistic plots in our 2D example.
		   	 	\item compare test sequence $\zeta$ with that of closest sequence: if they are both less than $\zeta_{low}$, mark as True Negative (for confabulation). If $\zeta>\zeta_{high}$, False Positive. Otherwise "unsure" (this lowers coverage).
				\end{enumerate}		   	   
		     
		 \item Negative examples
		        \begin{enumerate}
		   	 \item extract pairs of sequential sentences from CFR sequences (possibly same as above)
		   	 	\item reword with LLM to change second sentence significantly; use LLM-as-judge with human spot-checking to keep only sequences that have significantly changed semantics;
		   	 		\begin{itemize}
		   	 		   \item: Opposite-type example
		   	 			\begin{itemize}
		   	 				\item: Generative Prompt: \textit{``Generate a new second sentence that expresses the opposite or a clear contradiction of the original second sentence. - Make the contradiction clear in the new second sentence without changing the first sentence."}. Note we are generating what may be `easy' negatives here in order to get a clear signal of the technique working --- this also gives optimistic results for the baselines.
		   	 			\item: Quality-check Prompt: \textit{``Accept only if:
- the rewritten second sentence is coherent with the first sentence
- it clearly expresses the opposite or a contradiction of the original second sentence
- it is not identical to the original second sentence
Reject if any of those are not true.'' }
		   	 			\end{itemize}
		   	 			\item: Warped-type example
		   	 			\begin{itemize}
		   	 				\item: Generative Prompt: \textit{``Generate a new second sentence that radically warps the meaning second sentence while keeping it related. - Make the warping clear in the new second sentence without changing the first sentence."}
		   	 			\item: Quality-check Prompt: \textit{``Accept only if:
- the rewritten second sentence is coherent with the first sentence
- it clearly expresses a warped version of the original second sentence
- it is not identical to the original second sentence
Reject if any of those are not true.'' }
		   	 			\end{itemize}
		   	 		\end{itemize}
		   	 	\item check $\zeta$ of each re-worded pair
		   	 	\item for each re-worded pair, find closest pair (by similarity with first sentence) in VSDB, and its $\zeta$; if $\zeta>\zeta_{high}$, consider this pair `too noisy'.
		   	 	\item compare test sequence $\zeta$ with that of its closest sequence: if they are both less than $\zeta_{low}$, mark as False Negative (for untruthfulness). If $\zeta>\zeta_{high}$,  True Positive.
				\end{enumerate}	
		\item Compute F1 Score
			\begin{itemize}
			\item P = TP/(TP + FP)
			\item R = TP/(TP + FN)
			\item F1 = 2P*R/(P + R)
			\end{itemize}
		\item Other metrics (MCC, AUC, Cross Entropy) follow standard procedure.
		\item We also compute Area under the Risk Curve (AURC), measuring the trade-off between error-rate and coverage (percent of examples not in the "unsure" bucket).
	\end{itemize}
\end{itemize}

Tables~\ref{tab:transf1_appendix},\ref{tab:transf2} show examples of our Positive and Negative generated examples.

\begin{table*}[t]
  \caption{Examples of Ground Truth sequences from CFR transformed with a LLM to very similar sequences, which are labeled as Negative for confabulation. The first two rows show typical examples from this procedure, while the last row shows a rejected example where the Ground Truth is in too noisy a region of concept embedding space. This is typical when the sentence concept is very simple and has many different continuations in the corpus.}
  \label{tab:transf1_appendix}
  \centering

\begin{tabularx}{\linewidth}{>{\raggedright\arraybackslash}X>{\raggedright\arraybackslash}Xcc}
\toprule
\textbf{Ground Truth Sequence} & \textbf{Transformed Positive Sequence} & $\zeta_{GT}$ & $\zeta_{transf}$ \\
\midrule
{}["A discussion of work to be performed during the next reporting period.","Between scheduled reporting dates the grantee(s) also shall immediately inform the Grant Officer's Technical Representative (GOTR) of significant developments affecting their ability to accomplish the work."] & ["A discussion of work to be performed during the next reporting period.", "Between scheduled reporting dates, the grantee(s) must also promptly notify the Grant Officer's Technical Representative (GOTR) of any significant developments that affect their ability to complete the work."] & $4.66 \cdot 10^{-6}$ & $1.04$ \\
{}["We believe that these proposals for enhanced local access to OCA information would help to realize the benefits of public disclosure of OCA information identified in the benefit assessment and would help satisfy the public's interest in access at the local level to information about the sources of chemical accident risks that could affect them directly.","We anticipate that members of the public seeking OCA information held by LEPCs and local fire departments would be more likely to ask about the other information available from LEPCs under EPCRA regarding chemical hazards in the community."] & ["We believe that these proposals for enhanced local access to OCA information would help to realize the benefits of public disclosure of OCA information identified in the benefit assessment and would help satisfy the public's interest in access at the local level to information about the sources of chemical accident risks that could affect them directly.","We anticipate that members of the public seeking OCA information held by LEPCs and local fire departments would be more likely to inquire about the other information available from LEPCs under EPCRA regarding chemical hazards in the community."] & $1.07 \cdot 10^{-5}$ & $0.48$ \\
{}["We see no special concerns warranting an exception to this policy.", "Based on these results we conclude that this waste does not pose risk to human health and the environment at levels that warrant listing."] & ["We see no special concerns warranting an exception to this policy.", "Based on these results, we conclude that this waste does not pose a risk to human health and the environment at levels that warrant listing."] & $3.59$ & $3.72$ \\
\bottomrule
\end{tabularx}

\end{table*}

\begin{table*}[t]
  \caption{Examples of Ground Truth sequences from CFR transformed with a LLM to very different sequences, labeled as Negative for groundedness. The first two rows show typical examples of simple negation from this procedure, while the last row shows a more warped transformation.}
  \label{tab:transf2}
  \centering

\begin{tabularx}{\linewidth}{>{\raggedright\arraybackslash}X>{\raggedright\arraybackslash}Xcc}
\toprule
\textbf{Ground Truth Sequence} & \textbf{Transformed Negative Sequence} & $\zeta_{GT}$ & $\zeta_{transf}$ \\
\midrule
{}["(7) Describe the language capabilities of staff proposed for this study.", "The proposal clearly describes the (1) qualifications, commitment, and epidemiologic skills and experience of the project director and his/her ability to devote adequate time and effort to provide effective leadership; (2) qualifications and experience of other staff involved in the project to accomplish the proposed activity, and their commitment and time they will devote; (3) successful experience the project director and staff have in managing, coordinating and conducting similar or related projects; (4) a study coordinator with epidemiologic training and experience who is able to devote at least 50 percent of his or her time to this project; and (5) facilities, space, and equipment necessary for conducting the project."] & ["(7) Describe the language capabilities of staff proposed for this study.","The proposal fails to describe any qualifications, commitment, or epidemiologic skills of the project director or staff, provides no evidence of their experience, does not identify a study coordinator with epidemiologic training, and omits information about the necessary facilities, space, or equipment."] & $4.29 \cdot 10^{-6}$ & $14.85$ \\
{}["The need to ensure an adequate supply of generation usually was met through requirements imposed by states on franchise utilities to build or buy adequate power resources to meet demand consistently.", "Today, however, in states such as California, the adequacy of local power resources depends, not just on state requirements, but also on whether market prices are sufficient to elicit adequate supplies, through construction or otherwise."] & ["The need to ensure an adequate supply of generation usually was met through requirements imposed by states on franchise utilities to build or buy adequate power resources to meet demand consistently.","Today, in states such as California, the adequacy of local power resources depends solely on state requirements, with market prices playing no role in securing sufficient supplies."] & $2.57 \cdot 10^{-5}$ & $5.05$ \\
{}["Because the Exchange administers the Pilot Fee Structure as part of its rules, the Commission requests that the Exchange provide within 45 calendar days a thorough description of each fee that is permissible under the Pilot Fee Structure.", "The description should clearly identify the circumstances in which a distribution intermediary may assess a particular fee."] & ["Because the Exchange administers the Pilot Fee Structure as part of its rules, the Commission requests that the Exchange provide within 45 calendar days a thorough description of each fee that is permissible under the Pilot Fee Structure.","The description must detail the mysterious, almost mythical circumstances that allow a distribution intermediary to conjure and impose a fee at will."] & $4.49 \cdot 10^{-8}$ & $6.14$ \\
\bottomrule
\end{tabularx}

\end{table*}

We chose 10 years of the CFR as a validation set: 2000, 2001, 2005, 2008, 2011, 2013, 2017, 2020, 2022, and 2025.
Choosing 1k Positive and 10k Negative example pairs from each year, we compared observed deltas with measurements of the Concept Field for that year, with hyperparameter choices for the number of nearest neighbors ($topN$), number of top components compared for $\zeta$ calculation ($topN_\zeta$), and maximum neighbor distance ($dist_{max}$). Below in Table \ref{tab:val_res}  is an example analysis for one set of hyperparameter choices (here we choose $\zeta_{low} = \zeta_{high} = 3$):

\begin{table*}[t]
  \caption{Year-by-year results for parameter choice (topN=10, $topN_\zeta$=10, $dist_{max}$=0.3) on the Validation Set.}
  \label{tab:val_res}
  \centering

\begin{tabularx}{\linewidth}{>{\raggedright\arraybackslash}p{0.18\linewidth}>{\centering\arraybackslash}X>{\centering\arraybackslash}X>{\centering\arraybackslash}X>{\centering\arraybackslash}X>{\centering\arraybackslash}X>{\centering\arraybackslash}X}
\toprule
Measure & 2000 & 2001 & 2005 & 2008 & 2011 & 2013 \\
\midrule
{}\textbf{TP} & 618 & 583 & 619 & 625 & 578 & 585 \\
{}\textbf{TN} & 885 & 872 & 846 & 835 & 833 & 835 \\
{}\textbf{FP} & 61 & 62 & 58 & 85 & 66 & 60 \\
{}\textbf{FN} & 138 & 174 & 174 & 128 & 159 & 161 \\
{}\textbf{P} & 0.91 & 0.90 & 0.91 & 0.88 & 0.90 & 0.91 \\
{}\textbf{R} & 0.82 & 0.77 & 0.78 & 0.83 & 0.78 & 0.78 \\
{}\textbf{F1} & 0.86 & 0.83 & 0.84 & 0.85 & 0.84 & 0.84 \\
{}\textbf{MCC} & 0.76 & 0.72 & 0.73 & 0.74 & 0.72 & 0.73 \\
{}\textbf{AUC} & 0.94 & 0.92 & 0.93 & 0.93 & 0.93 & 0.93 \\
{}\textbf{LogLoss} & 1.37 & 1.41 & 1.35 & 1.55 & 1.45 & 1.32 \\
\bottomrule
\end{tabularx}

\begin{tabularx}{\linewidth}{>{\raggedright\arraybackslash}p{0.18\linewidth}>{\centering\arraybackslash}X>{\centering\arraybackslash}X>{\centering\arraybackslash}X>{\centering\arraybackslash}X>{\centering\arraybackslash}X}
\toprule
Measure & 2017 & 2020 & 2022 & 2025 & \textbf{Total} \\
\midrule
{}\textbf{TP} & 605 & 495 & 572 & 585 & \textbf{5865} \\
{}\textbf{TN} & 855 & 824 & 793 & 828 & \textbf{8406} \\
{}\textbf{FP} & 71 & 68 & 89 & 81 & \textbf{701} \\
{}\textbf{FN} & 163 & 134 & 149 & 148 & \textbf{1528} \\
{}\textbf{P} & 0.89 & 0.88 & 0.87 & 0.88 & \textbf{0.89} \\
{}\textbf{R} & 0.79 & 0.79 & 0.79 & 0.80 & \textbf{0.79} \\
{}\textbf{F1} & 0.84 & 0.83 & 0.83 & 0.84 & \textbf{0.84} \\
{}\textbf{MCC} & 0.72 & 0.72 & 0.70 & 0.72 & \textbf{0.73} \\
{}\textbf{AUC} & 0.93 & 0.93 & 0.93 & 0.93 & \textbf{0.93} \\
{}\textbf{LogLoss} & 1.41 & 1.57 & 1.46 & 1.45 & \textbf{1.43} \\
\bottomrule
\end{tabularx}

\end{table*}

As we can see there is little variation between the performance on any given year, showing the stability of the technique.
Ablation analysis over suitable combinations of these hyperparameters yielded the following results in Table \ref{tab:val_ab_cfr} compiled over all years of the validation set.

\begin{table*}[t]
  \caption{Ablation across parameters of the IDW averaging scheme on the Validation Set for groundedness in the CFR. Here only sentence pairs with $\zeta$ less than 0.5 or greater than 3 are accepted, all others being too "unsure."  The inverse power $p$ was set to 1 for all tests as preliminary analysis indicated essentially no difference from $p>1$.}
  \label{tab:val_ab_cfr}
  \centering

\begin{tabularx}{\linewidth}{>{\raggedright\arraybackslash}p{0.18\linewidth}>{\centering\arraybackslash}X>{\centering\arraybackslash}X>{\centering\arraybackslash}X>{\centering\arraybackslash}X>{\centering\arraybackslash}X>{\centering\arraybackslash}X}
\toprule
Measure & Case 1 & Case 2 & Case 3 & Case 4 & Case 5 & Case 6 \\
\midrule
{}$topN$ & 5 & 10 & 50 & 10 & 10 & 10 \\
{}$d_{max}$ & 0.1 & 0.3 & 0.5 & 0.3 & 0.3 & 0.3 \\
{}$topN_\zeta$ & 10 & 10 & 10 & 3 & 50 & 1024 \\
{}explanation & small neighborhood & defaults & larger neighborhood & reduced sensitivity & medium sensitivity & full sensitivity \\
{}TP & 558 & 4946 & 4953 & 4560 & 5087 & 5095 \\
{}TN & 359 & 2645 & 2654 & 3246 & 2438 & 2366 \\
{}FP & 33 & 238 & 235 & 434 & 124 & 120 \\
{}FN & 8 & 84 & 87 & 258 & 55 & 58 \\
{}P & 0.944 & 0.954 & 0.955 & 0.913 & 0.976 & 0.977 \\
{}R & 0.986 & 0.983 & 0.983 & 0.946 & 0.989 & 0.989 \\
{}F1 & 0.965 & 0.968 & 0.969 & 0.929 & \textbf{0.983} & 0.983 \\
{}MCC & 0.912 & 0.911 & 0.912 & 0.833 & \textbf{0.948} & 0.947 \\
{}AUC & 0.975 & 0.979 & 0.979 & 0.949 & \textbf{0.989} & 0.989 \\
{}CE Loss & 0.804 & 0.680 & 0.675 & 1.150 & 0.459 & \textbf{0.440} \\
\bottomrule
\end{tabularx}

\end{table*}

Beyond the simple VDB baseline described in the main text, for the final evaluation comparing to a VDB baseline, we drew an equal stratified sample of 160 examples across the hidden initial GPT-OSS-120B support classes. Three human reviewers independently read the CFR context, question, and GPT-OSS-20B answer while blinded to all model labels and scores. Twenty rows were shared across reviewers; seven disagreements were adjudicated while the model outputs remained hidden. The final labels were 41 supported, 28 partially supported, 27 unsupported, and 64 contradicted, with Fleiss' $\kappa=0.678$ on the overlapping reviews. The stratification supports comparison across categories but does not estimate their prevalence in the full CFR population.

For this experiment, the scored transition is explicitly $s_1=$ question and $s_2=$ GPT-OSS-20B answer. Thus the anchor is a generated question rather than a naturally adjacent CFR declarative sentence. This makes the experiment a realistic stress test, but introduces a distribution shift relative to the transitions used to construct the field.

We evaluate a binary selective endpoint in which supported answers are negative and unsupported or contradicted answers are positive; partially supported answers are excluded. For $\zeta$ and the retrieval-cosine baseline, lower and upper thresholds were selected on four folds and evaluated on the held-out fold in five-fold stratified cross-validation. For the direct retrieval baseline, we built a separate normalized cosine VDB over all 1,470,836 adjacent sentence pairs in the 2016 CFR. Each GPT-OSS-20B question--answer pair is embedded as one query and retrieves its global top-1 sentence-pair match; cosine distance to that retrieved pair is the baseline score. This does not alter the original GPT-OSS-20B answer. GPT-OSS-120B is evaluated as a categorical support judge on the same human-labeled examples, with partially supported treated as abstention in the binary selective calculation.

\begin{table*}[t]
  \caption{Human-validated CFR QA comparison on the same 160 examples. The $\zeta$ and direct pair-VDB cosine rows use five-fold held-out cross-validation; GPT-OSS-120B emits a categorical decision, so AUROC is not applicable.}
  \label{tab:cfr-qa-human-comparison}
  \centering

  \begin{tabularx}{\linewidth}{>{\raggedright\arraybackslash}p{0.18\linewidth}>{\centering\arraybackslash}X>{\centering\arraybackslash}X>{\centering\arraybackslash}X>{\centering\arraybackslash}X>{\centering\arraybackslash}X}
\toprule
Measure & $\zeta$ & Pair-VDB cosine (top-1) & $\zeta$ & Pair-VDB cosine (top-1) & GPT-OSS-120B judge \\
\midrule
{}Operating point & 50\% target & 50\% target & 84\% target & 84\% target & Fixed selective decision \\
{}Coverage & 0.500 & 0.523 & 0.795 & 0.848 & 0.841 \\
{}$F_1$ & 0.850 & 0.857 & 0.838 & 0.844 & 0.993 \\
{}Specificity & 0.056 & 0.062 & 0.033 & 0.034 & 0.972 \\
{}Balanced accuracy & 0.528 & 0.512 & 0.517 & 0.505 & 0.986 \\
{}MCC & 0.203 & 0.051 & 0.155 & 0.028 & 0.980 \\
{}AUROC & 0.559 & 0.685 & 0.559 & 0.685 & -- \\
\bottomrule
\end{tabularx}

\end{table*}

Table~\ref{tab:cfr-qa-human-comparison} reports observed held-out cross-validation estimates. Stratified bootstrap intervals give AUROC 0.559 [0.440, 0.672] for $\zeta$ and 0.685 [0.588, 0.781] for direct top-1 pair-VDB cosine. The paired cosine-minus-$\zeta$ AUROC difference is $0.126$ with 95\% CI $[0.0003, 0.255]$. This interval is only marginally above zero, so it supports a higher ranking AUROC for this cosine baseline on this sample but not a broad superiority claim. Both score-based methods have low held-out specificity at the evaluated coverages, and their $F_1$ values are consequently driven primarily by recall on the hallucinated class. They should not be interpreted as strong standalone factual-confabulation detectors. GPT-OSS-120B is substantially stronger as a direct semantic judge on this sample: it achieves selective $F_1=0.993$, specificity 0.972, balanced accuracy 0.986, and MCC 0.980 at 84.1\% coverage. Its exact four-class accuracy is 0.688 and macro-$F_1$ is 0.675.

\subsection*{G. Creative departure from the Gutenberg corpus}
\label{subsec:create}

\begin{itemize}
\item Experimental Procedure
	\begin{itemize}
		  \item as creativity assistant, essential difference from confabulation: we are looking for confirmation that an idea flow is \textit{not} in the corpus, something LLM cannot do
		   \item Negative examples
		   	 \begin{enumerate}
		   	 	\item extract pairs of sequential sentences from Gutenberg sequences
		   	 	\item reword with LLM to change \textbf{both sentences }slightly; use LLM-as-judge with human spot-checking to keep only pairs that do not significantly change semantics;
		   	 	   \begin{itemize}
		   	 			\item: Generative Prompt: \textit{``You will be given a sentence pair.
Rewrite both sentences with slight wording changes while preserving their meanings.
Requirements:
- Preserve the meaning of the first sentence.
- Preserve the meaning of the second sentence.
- Make only slight wording changes rather than broad paraphrases.
- Do not leave either sentence identical to the original."}
		   	 			\item: Quality-check Prompt: \textit{``Accept only if:
- the rewritten first sentence preserves the meaning of the original first sentence
- the rewritten second sentence preserves the meaning of the original second sentence
- both rewritten sentences remain coherent with each other as a pair
- both are only slight wording changes rather than broad paraphrases
- neither rewritten sentence is identical to its original sentence
Reject if any of those are not true.'' }
		   	 		\end{itemize}
		   	 	\item check $\zeta$  of each re-worded pair: if 'out of corpus' or 'insufficient statistics', \textbf{mark as False Positive}.
		   	 	\item for re-worded pair, find closest pair (by similarity with first sentence) in VSDB, and its $\zeta$ ; if $\zeta>\zeta_{high}$, consider this pair "too noisy"
		   	 	\item compare test sequence $\zeta$  with closest sequence: if they are both less than $\zeta_{low}$, mark as True Negative (for novelty). if $\zeta>\zeta_{high}$, False Positive.
				\end{enumerate}		   	   
		     
		 \item Positive examples
		        \begin{enumerate}
		   	 \item extract pairs of sequential sentences from Gutenberg sequences (possibly same as above)
		   	 	\item reword with LLM to\textbf{ change first sentence slightly and second sentence radically}; use LLM-as-judge with human spot-checking to keep only sequences that have significantly changed semantics;
		   	 		 \begin{itemize}
		   	 			\item: Generative Prompt: \textit{``You will be given a sentence pair.
Rewrite both sentences.
Requirements:
- Rewrite the first sentence with only a slight wording change while preserving its meaning. Do not leave it the same as the original.
- Generate a new second sentence that radically warps the meaning."}
		   	 			\item: Quality-check Prompt: \textit{``Accept only if:
- the rewritten first sentence preserves the meaning of the original first sentence
- the rewritten first sentence is only a slight wording change rather than a broad paraphrase
- the rewritten sentence is different from the original first sentence
Reject if any of those are not true.'' }
		   	 		\end{itemize}
		   	 	\item check $\zeta$  of each re-worded pair: if 'out of corpus' or 'insufficient statistics', \textbf{mark as True Positive}. 
		   	 	\item for re-worded pair, find closest pair (by similarity with first sentence) in VSDB, and its $\zeta$ ; if $\zeta>\zeta_{high}$, consider this pair "too noisy"
		   	 	\item compare test sequence $\zeta$  with closest sequence: if they are both less than $\zeta_{low}$, mark as False Negative (for novelty). if $\zeta>\zeta_{high}$, True Positive.
				\end{enumerate}	
		\item Compute F1 Score
			\begin{itemize}
			\item P = TP/(TP + FP)
			\item R = TP/(TP + FN)
			\item F1 = 2P*R/(P + R)
			\end{itemize}
		\item Other metrics (MCC, AUC, Cross Entropy) follow standard procedure as before.
	\end{itemize}
\end{itemize}

As for creativity detection, we generate Negative examples by lightly adjusting the ground truth sequences, as shown below in Table \ref{tab:transf3}.

\begin{table*}[t]
  \caption{Examples of Ground Truth sequences from the Gutenberg Corpus transformed with a LLM to very similar sequences, labeled as Negative examples for creativity. The first two rows show typical examples from this procedure, while the last row shows a rejected example where the Ground Truth is in too noisy a region of embedding space.}
  \label{tab:transf3}
  \centering

\begin{tabularx}{\linewidth}{>{\raggedright\arraybackslash}X>{\raggedright\arraybackslash}Xcc}
\toprule
\textbf{Ground Truth Sequence} & \textbf{Transformed Negative Sequence} & $\zeta_{GT}$ & $\zeta_{transf}$ \\
\midrule
{}["In returning now to the adventures of Siegfried there is little more to be described except the finale of an opera.",Siegfried, having passed unharmed through the fire, wakes Brynhild and goes through all the fancies and ecstasies of love at first sight in a duet which ends with an apostrophe to 'leuchtende Liebe, lachender Tod!', which has been romantically translated into 'Love that illumines, laughing at Death,' whereas it really identifies enlightening love and laughing death as involving each other so closely as to be usually one and the same thing."] & ["Returning now to Siegfried's adventures, there is little else to describe beyond the opera's finale.", "Siegfried, having emerged unscathed from the flames, awakens Brynhild and experiences the whims and raptures of love at first sight in a duet that concludes with an apostrophe to 'leuchtende Liebe, lachender Tod!', romantically rendered as \"Love that illumines, laughing at Death,' though it actually denotes that enlightened love and laughing death are so intertwined that they are often regarded as the same."] & $1.95\cdot 10^{-5}$ & 0.618 \\
{}["While teachers may be real authorities in subject-matter, they can never be anything more than assistants in the self-development of their students.","They should more openly assume this subordinate position, placing the primary responsibility upon the learner; they would then be less likely to subordinate the inner growth of the student, which it is their highest function to aid, to the mere
acquisition of knowledge."] & ["Although teachers can be genuine experts in their subject, they are at most helpers in their students' self-development.", "They ought to more openly adopt this subordinate role, putting the main responsibility on the learner; then they would be less inclined to subordinate the student's inner growth—its highest purpose to support - to mere knowledge acquisition."] & $4.21 \cdot 10^{-5}$ & 1.08 \\
{}["The moments passed.", "I heard the whistle of the approaching train."] & ["The moments went by.", "I heard the whistle from the approaching train."] & 6.20 & 1.45 \\
\bottomrule
\end{tabularx}

\end{table*}

For creativity detection, we generate Positive examples by radically transforming the second sentence, as shown below in Table \ref{tab:transf4}.

\begin{table*}[t]
  \caption{Examples of Ground Truth sequences from the Gutenberg Corpus transformed with a LLM to very different sequences, labeled as Positive for creativity.}
  \label{tab:transf4}
  \centering

\begin{tabularx}{\linewidth}{>{\raggedright\arraybackslash}X>{\raggedright\arraybackslash}Xcc}
\toprule
\textbf{Ground Truth Sequence} & \textbf{Transformed Positive Sequence} & $\zeta_{GT}$ & $\zeta_{transf}$ \\
\midrule
{}["There are some things that may not be discussed directly, and the conduct of life at a modern university--which is 
a reflection of life in the greater world--is one of these.", "Perry Blackwood and Ham did most of the talking, while Ralph, characteristically, lay at full length on the window-seat, interrupting with an occasional terse and cynical remark very much to the point."] & ["Behind the polished façade of campus life, hidden truths fester in silence, turning the modern university into a clandestine arena where society's unspoken experiments unfold.","Perry Blackwood and Ham did most of the talking, while Ralph, characteristically, lay at full length on the window-seat, interrupting with an occasional terse and cynical remark very much to the point."] & $4.21 \cdot 10^{-5}$ & $3.14$ \\
{}["The most popular works of fiction, such as leave nothing to our imagination.", "And to this craving after prose, who would not be lenient, that has at all known life, with its usual predominance of our lower and less courageous selves, our constant hankering after the cosey closed door and line of least resistance?"] & ["Even the most obscure, censored tomes insist on dictating every detail of our thoughts.", "And to this craving after prose, who would not be lenient, that has at all known life, with its usual predominance of our lower and less courageous selves, our constant hankering after the cosey closed door and line of least resistance?"] & $5.07 \cdot 10^{-5}$ & 4.59 \\
\bottomrule
\end{tabularx}

\end{table*}

An ablation analysis similar to that for the CFR above yields similar conclusions:

\begin{table*}[t]
  \caption{Ablation across parameters of the IDW averaging scheme on the Validation Set for groundedness in the Gutenberg Corpus. Here only sentence pairs with $\zeta$ less than 1.0 or greater than 1.5 are accepted, all others being too "unsure."  Here we see the optimal parameter values align with the CFR ablation analysis.}
  \label{tab:val_ab_gutenberg}
  \centering

\begin{tabularx}{\linewidth}{>{\raggedright\arraybackslash}p{0.18\linewidth}>{\centering\arraybackslash}X>{\centering\arraybackslash}X>{\centering\arraybackslash}X}
\toprule
Measure & Case 1 & Case 2 & Case 3 \\
\midrule
{}$topN$ & 10 & 10 & 10 \\
{}$d_{max}$ & 0.3 & 0.3 & 0.3 \\
{}$\zeta_{topN}$ & 10 & 50 & 1024 \\
{}explanation & reduced sensitivity & medium sensitivity & full sensitivity \\
{}TP & 534 & 570 & 179 \\
{}TN & 114 & 83 & 170 \\
{}FP & 90 & 67 & 60 \\
{}FN & 14 & 0 & 13 \\
{}P & 0.856 & 0.895 & 0.749 \\
{}R & 0.974 & 1.00 & 0.932 \\
{}F1 & 0.911 & \textbf{0.944} & 0.831 \\
{}MCC & 0.631 & \textbf{0.704} & 0.675 \\
{}AUC & 0.888 & \textbf{0.947} & 0.874 \\
{}CE Loss & 0.542 & \textbf{0.367} & 0.762 \\
\bottomrule
\end{tabularx}

\end{table*}

For a more realistic test, we chose 10 books known to be included in a 1000-book subset of our test VSDB, and found publicly-available translations of these into several non-English languages, as shown in Table~\ref{tab:neg_gutenberg} below:

\begin{table*}[t]
  \caption{Negative Novelty examples: we take the original English versions and find publicly-available translations into the Target Language.}
  \label{tab:neg_gutenberg}
  \centering

\begin{tabularx}{\linewidth}{>{\raggedright\arraybackslash}X>{\raggedright\arraybackslash}Xl}
\toprule
Book Title & Author & Target Language \\
\midrule
{}Père Goriot & Honoré de Balzac & French \\
{}The First Men in the Moon & H.G. Wells & French \\
{}Pride and Prejudice & Jane Austen & French \\
{}Illustrious Gaudissart & Honoré de Balzac & French \\
{}Pollyanna Grows Up & Eleanor H. Porter & Russian \\
{}Ann Veronica & H.G. Wells & Russian \\
{}The First Men in the Moon & H.G. Wells & Russian \\
{}The First Men in the Moon & H.G. Wells & Spanish \\
{}Moors and Christians & Pedro Antonio de Alarcón & Spanish \\
{}O Pioneers! & Willa Cather & Spanish \\
\bottomrule
\end{tabularx}

\end{table*}

For positive examples, we chose recently-added books known to be not in our scan of the Gutenberg Corpus (which had a cut-off of March 2026), as shown in Table~\ref{tab:pos_gutenberg} below:

\begin{table*}[t]
  \caption{Positive Novelty examples: we take English books added to the Gutenberg Corpus in May 2026, after our version which cut off in March 2026.}
  \label{tab:pos_gutenberg}
  \centering

\begin{tabularx}{\linewidth}{>{\raggedright\arraybackslash}X>{\raggedright\arraybackslash}Xl}
\toprule
Book Title & Author & Target Language \\
\midrule
{}The type-writer girl & Olive Pratt Rayner & English \\
{}The adventures of Pony Dexter & Harriet A. Cheever & English \\
{}Magpie—diplomat & W. C. Tuttle & English \\
{}The lily of Mordaunt & Mrs. Georgie Sheldon & English \\
{}The spark of Skeeter Bill & W. C. Tuttle & English \\
{}The trail of the White Knight & Bruce Graeme & English \\
{}The bungalow mystery & Carolyn Keene & English \\
{}Out of the depths or, The Rector's trial & Madeline Leslie & English \\
{}Such a happy day & Catharine Shaw & English \\
{}The burning of Chelsea & Walter Merriam Pratt & English \\
\bottomrule
\end{tabularx}

\end{table*}

For each work we chose 100 random sentence pairs, zeta-scored against our 1000-book Gutenberg VSDB, as well as cosine-scored against a VDB built on all sentences using a closest-first-sentence approximation and taking cosine-distance between full pairs  (in the case where the closest sentence in the VDB was the terminal sentence of a sequence, this distance is fixed at its maximal value of 1.). Table~\ref{tab:books_gutenberg_appendix} below summarizes our findings.

\begin{table*}[t]
  \caption{Scores on Selected Books: $\zeta$-scored VSDB vs. $cosine$-scored VDB}
  \label{tab:books_gutenberg_appendix}
  \centering

\begin{tabularx}{\linewidth}{>{\raggedright\arraybackslash}Xlrrrr}
\toprule
Book Title & Label & $\zeta$ & cosine-distance & $\zeta_{>2.4}$ & $cosdist_{>0.4}$ \\
\midrule
{}The type-writer girl & Positive & 2.76 & 0.569 & 0.36 & 0.66 \\
{}The adventures of Pony Dexter & Positive & 2.37 & 0.556 & 0.25 & 0.70 \\
{}Magpie—diplomat & Positive & 2.69 & 0.565 & 0.36 & 0.63 \\
{}The lily of Mordaunt & Positive & 2.51 & 0.558 & 0.39 & 0.54 \\
{}The spark of Skeeter Bill & Positive & 2.25 & 0.532 & 0.28 & 0.52 \\
{}The trail of the White Knight & Positive & 2.20 & 0.571 & 0.26 & 0.62 \\
{}The bungalow mystery & Positive & 2.02 & 0.578 & 0.16 & 0.53 \\
{}Out of the depths or, The Rector's trial & Positive & 2.65 & 0.517 & 0.30 & 0.55 \\
{}Such a happy day & Positive & 2.43 & 0.588 & 0.26 & 0.65 \\
{}The burning of Chelsea & Positive & 2.45 & 0.495 & 0.27 & 0.82 \\
{}Père Goriot & Negative & 1.98 & 0.376 & 0.14 & 0.35 \\
{}The First Men in the Moon (Fr) & Negative & 1.72 & 0.494 & 0.06 & 0.06 \\
{}Pride and Prejudice & Negative & 2.06 & 0.557 & 0.12 & 0.42 \\
{}Illustrious Gaudissart & Negative & 1.95 & 0.542 & 0.20 & 0.74 \\
{}Pollyanna Grows Up & Negative & 1.48 & 0.460 & 0.31 & 0.58 \\
{}Ann Veronica & Negative & 1.59 & 0.442 & 0.16 & 0.40 \\
{}The First Men in the Moon (Ru) & Negative & 1.64 & 0.472 & 0.07 & 0.36 \\
{}The First Men in the Moon (Sp) & Negative & 2.31 & 0.300 & 0.09 & 0.26 \\
{}Moors and Christians & Negative & 2.48 & 0.524 & 0.13 & 0.55 \\
{}O Pioneers! & Negative & 1.51 & 0.400 & 0.04 & 0.31 \\
\bottomrule
\end{tabularx}

\end{table*}

\begin{figure}[ht]
  \centering
  \begin{subfigure}[t]{0.48\linewidth}
    \includegraphics[width=\linewidth,height=0.22\textheight,keepaspectratio]{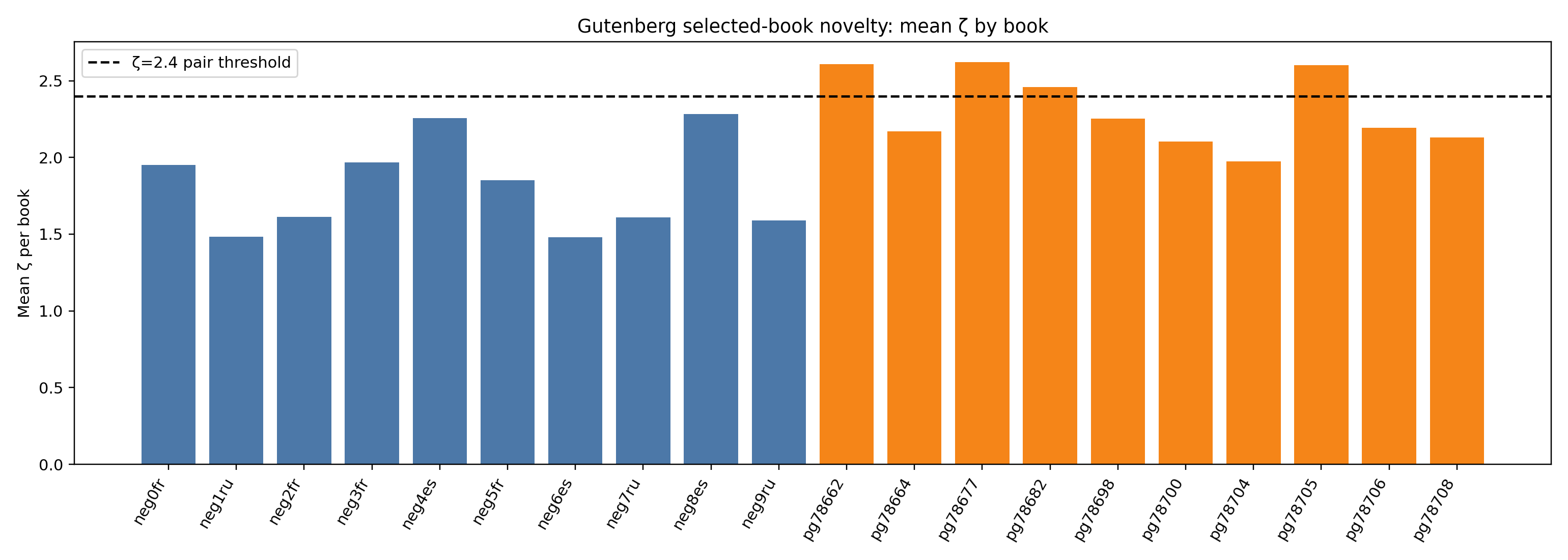}
    \caption{Mean $\zeta$ per book, by class.}
    \label{fig:meanzeta}
  \end{subfigure}
  \hfill
  \begin{subfigure}[t]{0.48\linewidth}
    \includegraphics[width=\linewidth,height=0.22\textheight,keepaspectratio]{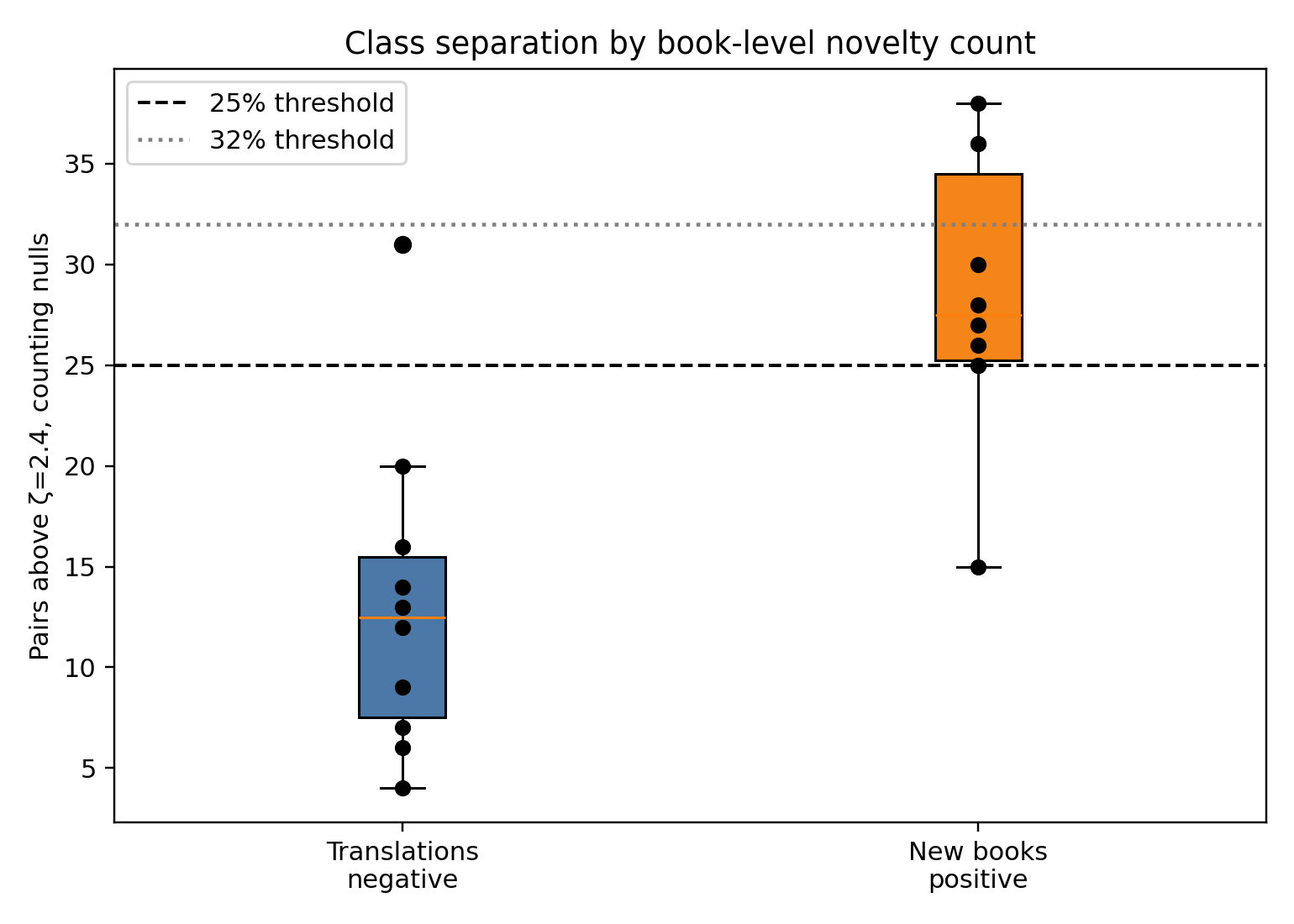}
    \caption{Distribution of pairs above $\zeta = 2.4$, by class.}
    \label{fig:boxcounts}
  \end{subfigure}
  \caption{Distribution-level views of the selected-book experiment. Left:
    mean $\zeta$ box plots show clear separation between positive and negative
    classes. Right: the distribution of above-threshold counts confirms that
    positive books cluster above the 25\,\% threshold (dashed) while most
    negative controls fall below it.}
  \label{fig:distrib}
\end{figure}

The positive and negative example books are fairly well-separated with $\zeta$ as a proxy novelty metric (see Fig.~\ref{fig:distrib}), and we determine that with $\zeta_{high}=2.4$ we obtain $R=1, P=0.69, F1=0.81$ for a VSDB-based method.

Meanwhile, a cosine sentence-pair index was constructed by extracting all adjacent
sentence pairs from the relevant 1000-book subsample of the Gutenberg corpus, embedding each pair
as a single SONAR text chunk, L2-normalising the resulting vectors, and
building a FAISS cosine index. The final index spans 2,421,383 sentence-pair
chunks.

At pair level, cosine distance was evaluated as a novelty signal (higher
distance indicates greater departure from the corpus). The pair-level results
over the 2,000 scored pairs (100 per book) are summarised in
Table~\ref{tab:cosine_pair}.

\begin{table*}[t]
  \caption{Pair-level cosine sentence-pair baseline performance.}
  \label{tab:cosine_pair}
  \centering

  \begin{tabularx}{\linewidth}{>{\raggedright\arraybackslash}Xrrrrr}
\toprule
\textbf{Method} & \textbf{AUC} & \textbf{Best $F_1$} & \textbf{Best threshold} & \textbf{AURC} & \textbf{Acc.\,@50\,\%} \\
\midrule
{}Cosine sentence-pair distance & 0.756 & 0.740 & 0.4215 & 0.264 & 0.741 \\
\bottomrule
\end{tabularx}
\end{table*}

The best pair-level operating point had $TP = 940$, $FP = 600$, $TN = 400$,
and $FN = 60$.

\subsubsection*{Book-Level Aggregation and Performance}

To enable a direct comparison with the $\zeta$ book-level decision rule, a
two-stage threshold was applied: first, a pair-distance threshold classifies
each individual pair as novel or not; second, a book is classified as novel if
the resulting count exceeds a book-level count threshold. Both thresholds were
selected jointly to maximise $F_1$ over the 20-book evaluation set.
Table~\ref{tab:cosine_book} reports the best in-sample configuration.

\begin{table*}[t]
  \caption{Book-level tuned cosine sentence-pair baseline performance.}
  \label{tab:cosine_book}
  \centering

  \begin{tabularx}{\linewidth}{>{\raggedright\arraybackslash}p{0.18\linewidth}>{\centering\arraybackslash}X}
\toprule
Measure & Case 1 \\
\midrule
{}\textbf{Pair dist.\ thresh.} & 0.2034 \\
{}\textbf{Book count thresh.} & 99 \\
{}TP & 10 \\
{}FP & 1 \\
{}TN & 9 \\
{}FN & 0 \\
{}$P$ & 0.909 \\
{}$R$ & 1.000 \\
{}$F_1$ & 0.952 \\
\bottomrule
\end{tabularx}
\end{table*}

\begin{figure}[ht]
  \centering
  \includegraphics[width=\linewidth,height=0.7\textheight,keepaspectratio]{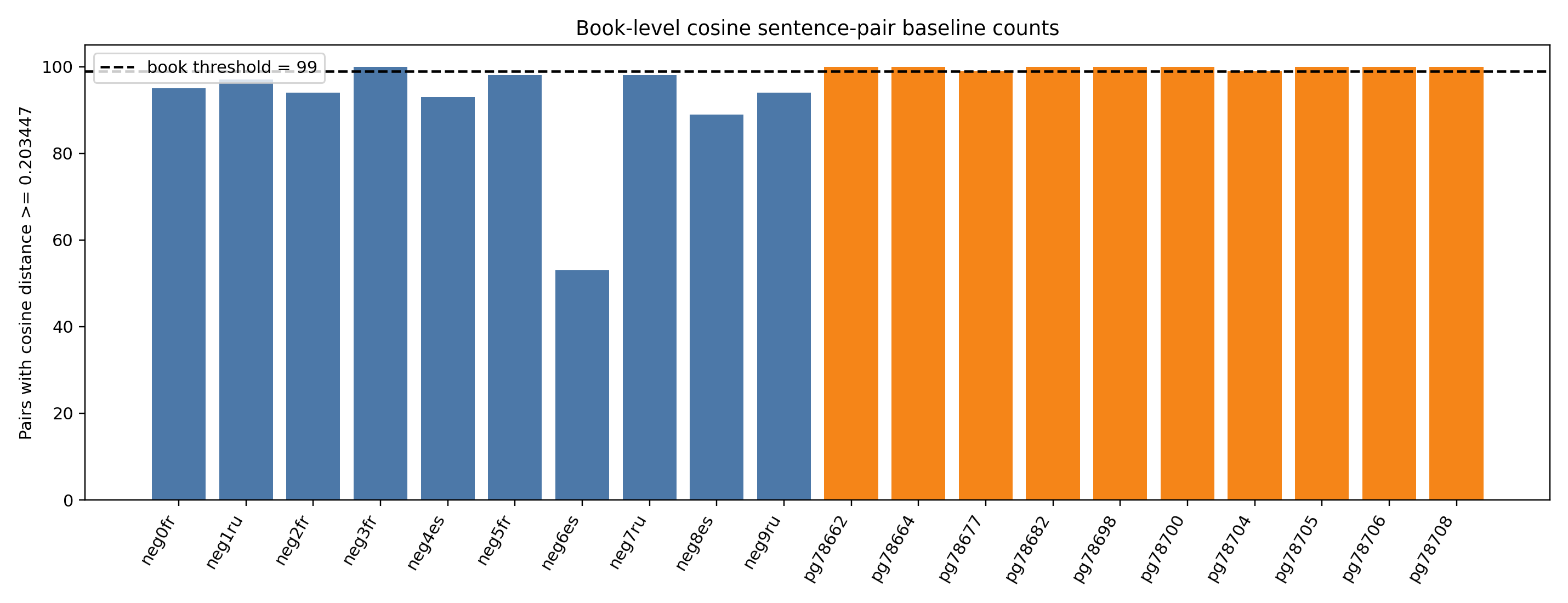}
  \caption{Book-level cosine baseline counts. A book is classified as novel if
    at least 99 of its 100 sampled pairs have nearest-neighbour cosine distance
    $\geq 0.2034$. The single false positive is \texttt{neg3fr}.}
  \label{fig:cosine_counts}
\end{figure}

On this 20-book benchmark, the tuned cosine baseline slightly outperforms the
$\zeta$ operating point at $\theta = 25\,\%$ ($F_1 = 0.952$ vs.\ $F_1 =
0.900$), however at optimized thresholds that could not be determined without prior calibration. A hallmark of this fact is that this threshold appears to be highly fine-tuned, for example, in Fig.~\ref{fig:cosine_counts}.
 The cosine result should therefore be treated as an optimistic
in-sample baseline, not as held-out generalisation performance.

\subsection*{H. Threshold Sensitivities on Corpora}

As mentioned in the main text, the Concept Field technique enjoys a certain stability across different corpora whereas other threshold-dependent techniques must be calibrated on a case-by-case basis.

\begin{table*}[t]
  \caption{Comparison of the Concept Field to various baselines with thresholds (lo,hi) tuned to give maximum performance at 50\% coverage of the original Gutenberg set of 9k+9k pos+neg test samples. Precision is defined on the ungrounded "confabulation" examples. AURC is computed over all possible lo/hi thresholds. Here we see the distance-based techniques enjoy a high performance on this artificial test set but with very different threshold values. Also the Concept Field technique captures many more Positive examples.}
  \label{tab:test_novelty_gut}
  \centering

\begin{tabularx}{\linewidth}{>{\raggedright\arraybackslash}p{0.18\linewidth}>{\centering\arraybackslash}X>{\centering\arraybackslash}X>{\centering\arraybackslash}X>{\centering\arraybackslash}X}
\toprule
Measure & VSDB + Field & VSDB + top1(L2) & VSDB + top1(cos) & VDB + top1(cos) \\
\midrule
{}lo & 1.2 & 0.08 & 0.01 & 0.02 \\
{}hi & 3.7 & 0.28 & 0.21 & 0.19 \\
{}TP & 4158 & 2449 & 2880 & 1068 \\
{}TN & 2366 & 5464 & 4505 & 6662 \\
{}FP & 332 & 2 & 3 & 0 \\
{}FN & 401 & 0 & 0 & 1 \\
{}P & 0.926 & 0.999 & 0.999 & 0.999 \\
{}R & 0.912 & 1.00 & 1.00 & 1.00 \\
{}F1 & 0.919 & 0.999 & 0.999 & 0.999 \\
{}MCC & 0.785 & 0.999 & 0.999 & 0.999 \\
{}AUC & 0.958 & 0.999 & 0.999 & 0.999 \\
{}AURC & 0.08 & 0.001 & 0.001 & 0.001 \\
\bottomrule
\end{tabularx}
\end{table*}

\subsection*{I. Divergence and Curl Analysis}

Representing a text corpus by a vector field,  we expect its geometric flow features, manifest in its divergence and curl, to have tangible semantic implications. Parallel to physical examples of vector fields in, say, fluids, having convergent regions (high absolute divergence and low curl), laminar regions (low divergence and curl) and vortices (high curl), in semantic embedding space regions of high divergence indicate some concept that all other logic flows tend to start at (logic sources) or end at (logic sinks); regions of low divergence and curl, implying the vector field is highly linear with its vector sequences flowing in the same direction, indicate `common reasoning'.  Regions of high curl, on the other hand, are semantically interesting as where vectors flow around some central concept, literally `beating around the bush': this may indicate avoidance regions in the corpus where certain topics are conspicuously missing --- for a LLM trained on the data in such a VSDB, it may also indicate where the LLM is likely to hallucinate for lack of training data.
 
High dimensions preclude us from systematically computing divergence $\overrightarrow{\triangledown} \cdot \overrightarrow{f}$ and curl $\overrightarrow{\nabla} \times \overrightarrow{f}$ of the Concept Field on a grid, but we can estimate these at the centroids of dense semantic clusters as follows: for divergence, we directly estimate
\begin{equation}
 D = \frac{d}{N}\sum_{i=1}^{N} \frac{ \overrightarrow{v_i}^T \overrightarrow{r_i}}{|r_i|^2}
\end{equation}
where $\overrightarrow{v_i}$ is the delta to the `next' vector of $\overrightarrow{p_i}$, the concept embedding of the i-th sentence in the cluster of size N, and $\overrightarrow{r_i} = \overrightarrow{p_i} - \overrightarrow{c}$ is the radial vector pointing from the centroid to the i-th concept embedding with $\overrightarrow{c}$ being the position of the cluster centroid; for curl, we compute the `Total Angular Momentum' of a cluster as
\begin{equation}
 \textbf{M} = \frac{d}{N}\sum_{i=1}^{N} \frac{  \overrightarrow{r_i} \overrightarrow{v_i}^T - \overrightarrow{v_i} \overrightarrow{r_i}^T  }{|r_i|^2}
\end{equation}
The Frobenius Norm $||\textbf{M}||$ of the matrix $M$ is then proportional to the total circulation about the centroid.

In this paper, these geometric findings are qualitative and hypothesis-generating. We find, for example, that regions of high negative divergence in the CFR are effective at exposing variations on government standard logic such as ``Written comments and information are requested and will be accepted on or before July 23, 2018.'' $\longrightarrow$ ``Interested persons are encouraged to submit comments using the Federal eRulemaking Portal at ...''.

 Examples: 
 \begin{itemize}
 \item  \textbf{high positive divergence (logic source) in English Premier League Commentary\cite{football}}
 	\begin{itemize}
 		\item group size: 20
 		\item $\sum(cos(t)/r^2)=-18.24$, $Avg=-18.24$   (although negative, this was `most positive')
 		\item centroid semantic: ``Can either side find a winner to the end?''
 		\item example sentences in cluster:
 			\begin{itemize}
 				   \item ``Can either side snatch a lead just ahead of half-time? '' $\longrightarrow$ EOS
         \item ``Can either side find a winning goal, or are we set for a stalemate?  '' $\longrightarrow$ EOS
        \item ``Can either side now find a late winning goal? '' $\longrightarrow$ EOS
 			\end{itemize}
 	\end{itemize}
 \item  \textbf{high negative divergence (logic sink) in English Premier League Commentary}
 	\begin{itemize}
 		\item group size: 283
 		\item $\sum(cos(t)/r^2)=-14679.69$, $Avg=-51.87$   (although negative, this was `most positive')
 		\item centroid semantic: ``GOOOOAAALL!! It is''
 		\item example sentences in cluster:
 			\begin{itemize}
 				   \item ``GOALLLLLL! '' $\longrightarrow$ ``WHAT A START FOR BURNLEY! 1-0!''
         \item ``GOOOOOOAAAAAAAALL!  '' $\longrightarrow$ ``NEWCASTLE TAKE THE LEAD THROUGH SCHAR!''
        \item ``WHAT A GOAAALLLLLLLLLLLLLLLL!  '' $\longrightarrow$ ``UNITED STRIKE FIRST!''
 			\end{itemize}
 	\end{itemize}
 \item  \textbf{high positive divergence (logic source) in CFR}
 	\begin{itemize}
 		\item group size: 10
 		\item $\sum(cos(t)/r^2)=-82.54$, $Avg=-11.79$
 		\item centroid semantic: ``The specification of complementarity field GPLD-1 in reference to the claimed geomagnetic performance standard TMP-07 includes an 8°C reference field limited in consideration of spatial telemetry reference magnitude of the relevant geomagnetic field; an 8°C reference field for the field, and an 8°C reference field for the field in consideration of the information and simulation variability of the reference field.''
 		\item example sentences in cluster:
 			\begin{itemize}
 				   \item ``The supplemental GMD event definition in proposed Reliability Standard TPL-007-2 contains a non-spatially-averaged reference peak geoelectric field amplitude component of 12 V/km, in contrast to the 8 V/km figure in the spatially-averaged benchmark GMD event definition. '' $\longrightarrow$ ``As NERC explains in its petition, the supplemental GMD event will be used to “represent conditions associated with localized enhancement of the geomagnetic field during a severe GMD event for use in assessing GMD impacts.”''
         \item ``With respect to the calculation of the reference peak geoelectric field amplitude component of the benchmark GMD event definition, Order No. 830 expressed concern with relying solely on spatial averaging in Reliability Standard TPL-007-1 because “the use of spatial averaging in this context is new, and thus there is a dearth of information or research regarding its application or appropriate scale.”'' $\longrightarrow$ ``NERC states that proposed Reliability Standard TPL-007-2 enhances currently-effective Reliability Standard TPL-007-1 by addressing reliability risks posed by GMDs more effectively and implementing the directives in Order No. 830.''
        \item ``The supplemental GMD event definition contains a higher, non-spatially-averaged reference peak geoelectric  field amplitude component than the benchmark GMD event definition (12 V/km versus  8 V/km). '' $\longrightarrow$ ``These three new requirements largely mirror existing Requirements R4, R5, and R6 that currently apply, and would continue to apply, only to benchmark GMD vulnerability and transformer thermal impact assessments.''
 			\end{itemize}
 	\end{itemize}
\item  \textbf{high negative divergence (logic sink) in CFR}
 	\begin{itemize}
 		\item group size: 6155
 		\item $\sum(cos(t)/r^2)=-17763.35$, $Avg=-34.43$
 		\item centroid semantic: ``Comments to submit are to be received by or before February 18, 2019.''
 		\item example sentences in cluster:
 			\begin{itemize}
 				   \item ``Written comments and information are requested and will be accepted on or before July 23, 2018.'' $\longrightarrow$ ``Interested persons are encouraged to submit comments using the Federal eRulemaking Portal at''
         \item ``Rebuttal comments should be submitted by October 26, 2018.'' $\longrightarrow$ ``Although there do not appear to be any issues relevant to approval or disapproval which would be facilitated by an oral presentation of views, data, and arguments, the Commission will consider, pursuant to Rule 19b-4, any request for an opportunity to make an oral presentation.''
        \item ``Comments and related material must be received by the Coast Guard on or before July 9, 2018.'' $\longrightarrow$ ``You may submit comments identified by docket number USCG-2018-0580 using the Federal eRulemaking Portal at ''
 			\end{itemize}
 	\end{itemize}
\item  \textbf{high curl (implicit topic) in Gutenberg}
\begin{itemize}
	\item group size: 708042
	\item $||\textbf{M}|| = 1095.98$ 
    \item centroid semantic: ``The Commission is aware of the difficulties encountered by the Member States in implementing the guidelines.''
    \item example sentences in cluster:
 			\begin{itemize}
 				   \item ``And now we will leave the earth and look at the pictures of airships and the wonderful planes, which, although improvements are made every day, are already capable of flying with an almost incredible speed and with a security which only a little while ago would have been considered quite impossible.'' $\longrightarrow$ ``Today air travel is not only a possibility but a realized fact, and it is difficult to realize that it has occurred in the last twenty years, and that before then the practical flying machines were unknown.''
         \item ``In these matters the Balance of Power is not less vital for international life and for the evolution of true cosmopolitan ideals than it is in mere Politics.'' $\longrightarrow$ ``And if we stand up in a battle for the lesser races it is not merely because they are small and need defence but because an element of right, a part of the civilization we intend to conquer, is with them and is part of their heritage.''
        \item ``Therefore, a negotiated solution for the final status of Jerusalem could be of a character different from that of the rest of the West Bank.'' $\longrightarrow$ ``The Gaza Strip is currently governed by Israeli military authorities and Israeli civil administration; it is American policy that the final status of the Gaza Strip will be determined by negotiations between the parties concerned; these negotiations will determine how this area should be governed.''
 			\end{itemize}
 	\end{itemize}
 	  	
 \end{itemize}

\begin{figure*}[htbp]
     \centering
     \begin{subfigure}[b]{0.31\linewidth}
         \centering
         \includegraphics[width=\linewidth,height=0.22\textheight,keepaspectratio]{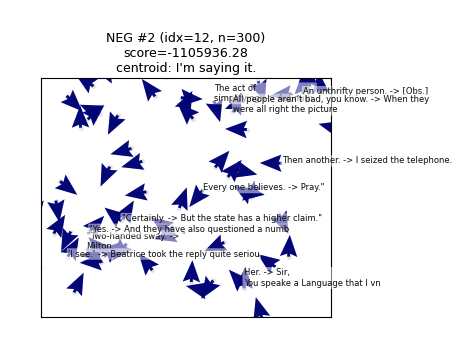}
         \caption{Part of a  highly negatively divergent region of the Gutenberg Corpus. }
         \label{fig:geom_a}
     \end{subfigure}
     \hfill 
     \begin{subfigure}[b]{0.31\linewidth}
         \centering
         \includegraphics[width=\linewidth,height=0.22\textheight,keepaspectratio]{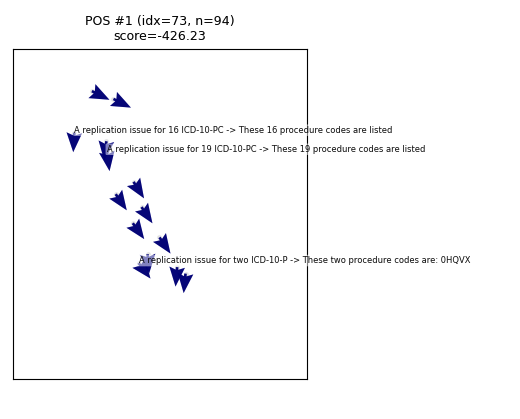}
         \caption{A region of the CFR where div and curl are minimal, hence indicating common logic flow.}
         \label{fig:geom_b}
     \end{subfigure}
     \hfill
     \begin{subfigure}[b]{0.31\linewidth}
         \centering
         \includegraphics[width=\linewidth,height=0.22\textheight,keepaspectratio]{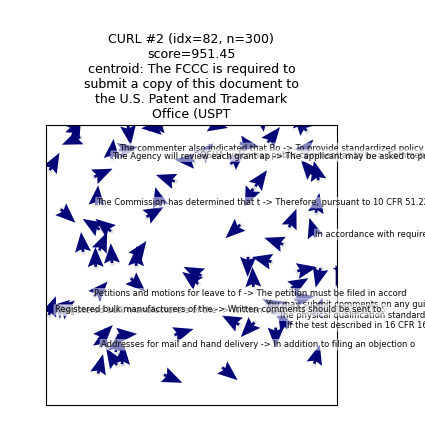}
         \caption{A region of the CFR with high curl.}
         \label{fig:geom_c}
     \end{subfigure}
     \caption{Examples of corpus regions in UMAP plots with extreme divergence and/or curl.}
     \label{fig:geom}
\end{figure*}


\end{document}